\RequirePackage{fix-cm}
\documentclass[]{ant_tech_report}

\microtypesetup{expansion=false}

\usepackage{longtable}          
\usepackage[edges]{forest}      
\usepackage[normalem]{ulem}     
\usepackage{textcomp}
\usepackage{url}

\IfFileExists{fontawesome5.sty}{\usepackage{fontawesome5}}{%
  }

\renewcommand{\eg}{\emph{e.g., }}
\newcommand{\etal}{\emph{et al.}\xspace}

\newcommand{\coord}[3]{(#1,\,#2,\,#3)}

\definecolor{whenColor}{RGB}{232,74,26}   
\definecolor{whereColor}{RGB}{47,107,47}  
\definecolor{howColor}{RGB}{106,27,154}   
\newcommand{\When}{\textcolor{whenColor}{\emph{When}}\xspace}
\newcommand{\Where}{\textcolor{whereColor}{\emph{Where}}\xspace}
\newcommand{\How}{\textcolor{howColor}{\emph{How}}\xspace}

\newcommand{\logoh}[2]{%
  \IfFileExists{#1}{\raisebox{-0.5\height}{\includegraphics[height=#2]{#1}}}{}}
\newcommand{\institutionlogos}{%
  \logoh{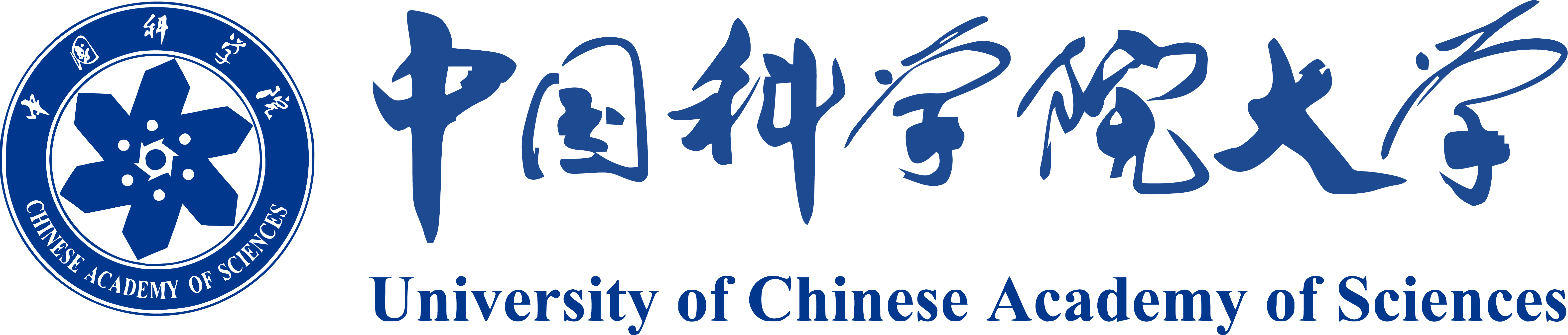}{7.5mm}\hspace{7mm}%
  \logoh{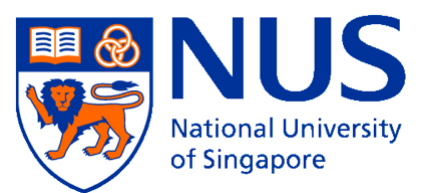}{8.5mm}\hspace{7mm}%
  \logoh{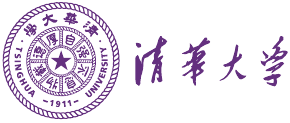}{9mm}%
}

\definecolor{accentblue}{HTML}{1F5FA9}    
\definecolor{accentbluebg}{HTML}{EEF4FB}  
\definecolor{accentpurple}{HTML}{6A4C93}
\definecolor{accentpurplebg}{HTML}{F4F0FA}

\colorlet{antblue}{accentblue}
\colorlet{metabg}{accentbluebg}

\hypersetup{linkcolor=antblue, citecolor=antblue, urlcolor=antblue}

\title{Continual Learning in Transition}

\author[1,6*]{Zhiyan Hou}
\author[2*\dag]{Dan Zhang}
\author[3*\dag]{Tao Feng}
\author[3]{Liyuan Wang}
\author[3]{Wei Li}
\author[1]{Xiangzhao Hao}

\authorbreak

\author[1]{Hongyan An}
\author[2]{Junfeng Fang}
\author[2]{Haokai Ma}
\author[4]{Zhaohui Xu}
\author[5]{Xinyu Tang}

\authorbreak

\author[1,6\dag]{Haiyun Guo}
\author[1,6,7]{Jinqiao Wang}
\author[2]{Tat-Seng Chua}

\affiliation[1]{Institute of Automation, Chinese Academy of Sciences}
\affiliation[2]{National University of Singapore}
\makeatletter
\g@addto@macro\affiliationlist{\\
  \affiliationformat[3]{Tsinghua University},
  \affiliationformat[4]{Alibaba Group}\\
  \affiliationformat[5]{evermind.ai}\\
  \affiliationformat[6]{School of Artificial Intelligence, University of Chinese Academy of Sciences}\\
  \affiliationformat[7]{Wuhan Artificial Intelligence Research Institute}}
\makeatother

\providecommand{\authorfootnotes}{%
  $^{*}$Equal contribution.\quad
  $^{\dag}$Corresponding authors.%
}

\fancypagestyle{firststyle}{%
  \fancyhf{}%
  \fancyhead[L]{%
    \raisebox{-7mm}[0pt][0pt]{%
      \vbox{\vskip 3mm\hbox{\institutionlogos}}%
    }%
  }%
  \fancyfoot[L]{%
    \raisebox{1.15\baselineskip}[0pt][0pt]{%
      \parbox[b]{\textwidth}{\rule{2in}{0.4pt}\\[3pt]\authorfootnotes}%
    }%
  }%
  \fancyfoot[C]{\thepage}%
}

\abstract{
Classical continual learning (CL) has primarily focused on enabling models to update and retain knowledge through parameter-centric mechanisms, e.g., training strategies, architectural designs, and weight adaptation. 
However, emerging paradigms are reshaping the scope of CL beyond this traditional model adaptation view. 
For instance, on-policy learning broadens the space of update mechanisms; test-time training extends CL from the training phase to inference; and external harness components such as memory, skill libraries, and interaction protocols extend the evolutionary boundaries of model capabilities far beyond the static parameter space. 
Collectively, these developments indicate a transition from parameter-centric learning toward system-level adaptation. 
To characterize this transition, we examine the evolution of continual learning through three dimensions: \emph{\When}, \emph{\How}, and \emph{\Where} learning occurs. 
The \emph{\How} dimension encompasses off-policy, on-policy, and beyond-gradient optimization mechanics. 
The \emph{\When} dimension captures evolution across pre-training, post-training, and inference-time stages. 
The \emph{\Where} dimension delineates updates occurring within internal parameters versus external structural constraints.
Anchored by this tri-axial framework, we systematically survey representative methods, trace the ongoing transition of continual learning, and discuss the key challenges, broader implications, and future directions arising from this paradigm shift.
}

\begin{document}
\maketitle


\section{Introduction}
\label{sec:intro}
Recent advances in large language models (LLMs) and Agentic AI~\cite{OpenAI2023GPT4TR, glm2024chatglm, bai2023qwen, deepseekai2025deepseekr1, team2026kimi, zeng2025glm, zeng2026glm}, including substantial progress on complex reasoning~\cite{zhang2024restmcts, zhang2025tdrm}, long-horizon task execution~\cite{team2026kimi, zeng2026glm}, and code generation~\cite{zhoubian2025rest, xia2024scenegenagent}, are widely regarded as a substantive step toward artificial general intelligence (AGI).
A genuinely general intelligence, however, must operate in open environments where the state of the world continuously evolves and task demands are constantly renewed.
This requires a system to update knowledge as the environment changes~\cite{wang2024memoryllm}, accumulate and recombine skills through interaction~\cite{wang2023voyager}, adjust behavioral policies in light of feedback~\cite{shinn2023reflexion}, and consolidate long-term memory across sessions and tasks~\cite{packer2023memgpt, zhong2024memorybank}.
Therefore, the path toward AGI lies not merely in training stronger static models, but in constructing agent systems that can continually improve after deployment~\cite{zhai2025agentevolver}. \textit{How to remain adaptive under distribution shift, persistently accumulate experience, and resist forgetting} is precisely what continual learning seeks to address~\cite{mccloskey1989catastrophic, khetarpal2022towards, wang2024comprehensivesurveycontinuallearning}.

\begin{figure*}[!t]
  \centering
  \includegraphics[width=0.95\linewidth]{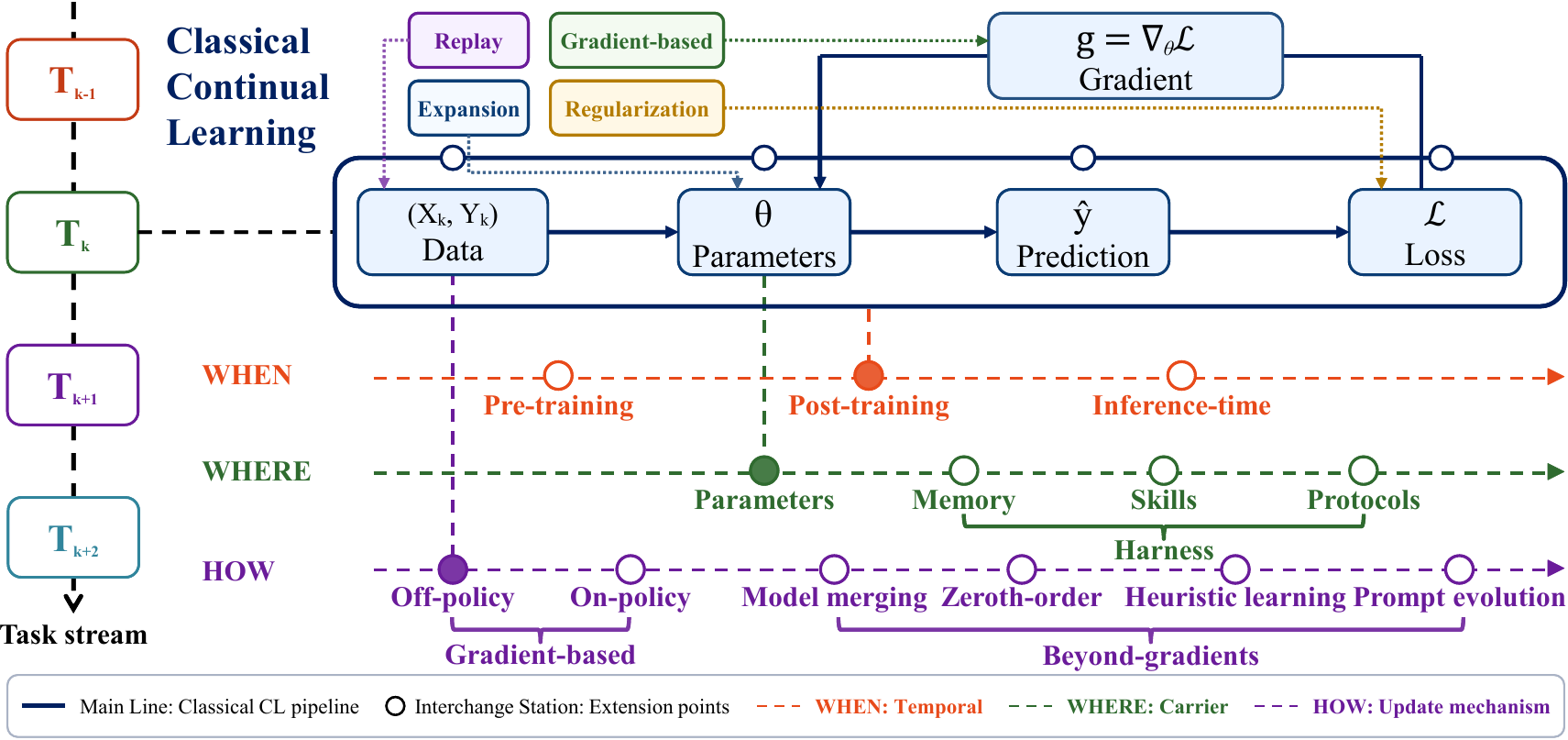}
  \vspace{-0.2cm}
  \caption{Overview of the three-dimensional view of continual learning developed in this survey. The main line (solid) depicts the classical continual learning pipeline, in which a stream of tasks is fitted by back-propagated gradient updates to model parameters, and circles mark the points at which classical method families intervene. The three dashed lines unfold the extensions examined in this survey: the \When line spans pre-training, post-training, and inference time; the \Where line spans parameters and the harness components (memory, skills, and protocols); and the \How line spans gradient-based updates (off-policy, on-policy) and beyond-gradient updates. Filled circles mark the position of the classical setting on each line.}
  \label{fig:overview_pipeline}
\end{figure*}

Classical continual learning (CL) has long centered on catastrophic forgetting and the stability--plasticity trade-off, with mainstream methods broadly grouped into four categories: \textit{replay-based} methods, \textit{gradient-based} methods, \textit{parameter-isolation or architecture-expansion} methods, and \textit{regularization-based} methods~\cite{wang2024comprehensivesurveycontinuallearning,wu2024continual} (Figure~\ref{fig:overview_pipeline}). 
With the rapid development of LLMs and agentic systems, however, the methodological landscape and application scenarios of CL are expanding rapidly, and the classical taxonomy can no longer fully capture these emerging developments.

The research landscape of CL is undergoing simultaneous changes along three interrelated dimensions (Figure~\ref{fig:overview_pipeline}). 
With respect to the \emph{learning mechanism} (the \How line in Figure~\ref{fig:overview_pipeline}), the change unfolds at two levels. 
Within \textit{gradient-based learning}, classical off-policy updates tend to cause the model to deviate substantially from its initial behavior distribution and thereby induce pronounced catastrophic forgetting, whereas on-policy paradigms, including reinforcement learning from human feedback (RLHF)~\cite{ouyang2022training} and reinforcement learning with verifiable rewards (RLVR)~\cite{shao2024deepseekmath,li2026unifying,yu2026knowrl,
hao2026clear}, exhibit notable advantages in mitigating forgetting~\cite{rlsRazor2025}.
This observation has subsequently inspired a line of on-policy post-training methods exemplified by On-Policy Distillation (OPD)~\cite{agarwal2024onpolicy} and, more recently, on-policy self-distillation (OPSD) for continual learning~\cite{shenfeld2026sdft}.
\textit{Beyond gradients}, approaches that do not rely on standard backpropagation, such as model merging~\cite{yang2024modelmerging}, zeroth-order optimization~\cite{malladi2023mezo}, heuristic learning~\cite{weng2026learning}, and prompt evolution~\cite{fernando2023promptbreeder}, have further broadened the design space of update mechanisms.
With respect to the \emph{learning timing} (the \When line in Figure~\ref{fig:overview_pipeline}), continual adaptation now spans the full model lifecycle, from continual pre-training and multi-stage post-training to the post-deployment inference loop.
Test-Time Adaptation (TTA)~\cite{wang2021tent} and Test-Time Training (TTT)~\cite{sun2024learning} introduce writable state or parameter updates at inference time, enabling models to perform on-the-fly adjustments when confronted with novel samples and distribution shifts.
With respect to the \emph{locus of capability} (the \Where line in Figure~\ref{fig:overview_pipeline}), the research focus has migrated from inside the model to the external harness layer that surrounds it~\cite{externalization2026}: stateless retrieval-augmented generation (RAG) has evolved into long-term memory systems with read--write semantics that persist across sessions and tasks~\cite{packer2023memgpt}; fixed tool sets have evolved into self-generated, reusable, and composable skill libraries~\cite{wang2023voyager}; and static interaction protocols and behavioral rules have evolved into adaptive protocols that are continually refined through reflection and feedback~\cite{shinn2023reflexion}. 
Capability accumulation and updates have thereby extended from the parameter space to multiple carriers situated outside the model.

\begin{figure*}[!t]
  \centering
  \includegraphics[width=\textwidth,keepaspectratio]{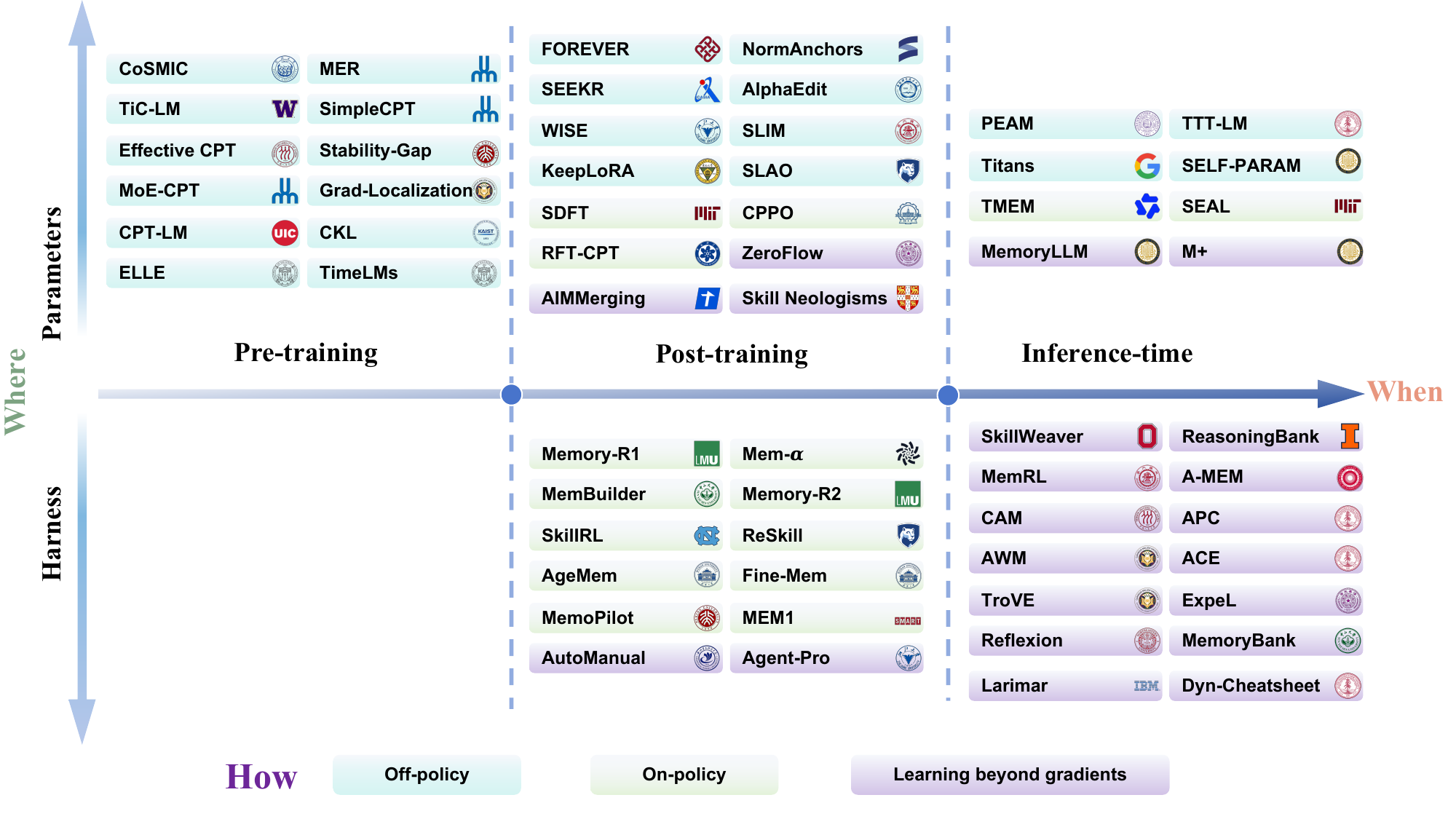}
  \vspace{-0.2cm}
  \caption{The three dimensions of continual learning with representative methods. Methods are placed by \emph{\When} capability evolution occurs, \emph{\Where} capability is accumulated, and \emph{\How} the update is driven. Selected abbreviated labels used in this figure and their corresponding references are listed in Table~\ref{tab:fig2-label-conventions}.}
  \label{fig:taxonomy_overview}
\end{figure*}

These developments collectively indicate that the research landscape of continual learning is undergoing concurrent changes along three dimensions: learning mechanism, learning timing, and locus of capability. 
The classical taxonomy, primarily organized around the canonical setting of training-time, parameter-level, off-policy gradient updates, can no longer naturally accommodate these emerging directions. 
Motivated by this observation, we reformulate continual learning in the era of large language models and agentic AI as \emph{continual capability evolution} across mechanism, timing, and locus, and propose a three-axis taxonomy that uniformly characterizes this process (Figure~\ref{fig:taxonomy_overview}). 
The \How axis characterizes the extension of update mechanisms from off-policy to on-policy and further to learning-beyond-gradients; the \When axis characterizes the extension of capability evolution across the lifecycle from pre-training and post-training to inference-time; and the \Where axis characterizes the extension of capability locus from parameters to the harness.

The main contributions of this survey are threefold.
\begin{itemize}[leftmargin=*,itemsep=0pt,parsep=0.5em,topsep=0.3em,partopsep=0.3em]
    \item We recast continual learning in the era of large language models and agentic AI as \emph{continual capability evolution}, viewed through three complementary questions: when capability evolves, where it is carried, and how it is updated. In contrast to existing LLM continual-learning surveys, which largely extend the classical parameter-centric categorization~\cite{wu2024continual, wang2024comprehensivesurveycontinuallearning}, this view brings the agent harness (memory, skills, and protocols) together with inference-time and gradient-free mechanisms into the same frame as parameter-level learning.

    \item Guided by this view, we review representative methods that are seldom examined together, spanning continual pre- and post-training, test-time training, reinforcement-learning-based alignment, model merging, memory systems, skill libraries, and prompt evolution, and locate each in a common \coord{\text{when}}{\text{where}}{\text{how}} space that makes their otherwise implicit relationships explicit.

    \item Reading this space as a whole, we characterize where methods concentrate and why, and identify the sparse and empty regions that mark concrete challenges and opportunities for continual learning in self-evolving agent systems.
\end{itemize}

\section{Revisiting Continual Learning}
\label{sec:revisiting}

Section~\ref{sec:intro} motivated the need to revisit the scope of continual learning in light of recent developments beyond the classical setting. Before the three dimensions can be developed, however, it is necessary to make explicit what the classical formulation held fixed. This section therefore revisits classical continual learning, including its formulation, objectives, and major method families (Section~\ref{sec:classical_cl}), and shows that they share three implicit assumptions. Section~\ref{sec:three_axis_extension} then turns these assumptions into the three questions of \emph{when}, \emph{where}, and \emph{how}, under which classical continual learning can be viewed as a particular point in a broader space.

\subsection{Classical Continual Learning and Its Three Implicit Assumptions}
\label{sec:classical_cl}

Classical continual learning studies a model that learns under a non-stationary data distribution. Training data arrive as a sequence of tasks $\mathcal{T}_1,\ldots,\mathcal{T}_K$, each associated with its own distribution $\mathcal{D}_k := p(X_k, Y_k)$ and loss $\mathcal{L}_k$~\cite{wu2024continual}. At step $k$, the model $\theta$ is updated while historical data $\{\mathcal{D}_i\}_{i<k}$ are inaccessible or available only in a restricted form. The ideal objective is to fit the current task while keeping the aggregate expected loss $\sum_{i\le k}\mathbb{E}_{(X, Y)\sim\mathcal{D}_i}[\mathcal{L}_i]$ over all observed tasks low, despite the fact that earlier data are no longer fully available. The central obstacle is \emph{catastrophic forgetting}~\cite{mccloskey1989catastrophic, goodfellow2013empirical}: updates that improve performance on $\mathcal{T}_k$ may perturb the representations or decision boundaries that support earlier tasks, sharply degrading their performance.

Classical methods are commonly evaluated against two coupled desiderata. The first is the \emph{plasticity and stability} trade-off: the model must remain plastic enough to acquire new tasks while maintaining stable knowledge of previous ones. From a Bayesian perspective, continual learning can be viewed as sequential posterior updating, where the posterior induced by previous tasks becomes the prior for the next task. This view motivates regularization-based formulations that approximate the preservation of previous knowledge by penalizing changes to parameters important for earlier tasks~\cite{wang2024comprehensivesurveycontinuallearning}. The second desideratum is \emph{intra-task and inter-task generalizability}: the learner should remain robust to the train/test gap within each task while adapting to distribution shifts across tasks, a requirement that continual reinforcement learning frames as retaining and transferring behavior across non-stationary environments~\cite{khetarpal2022towards}.

Around these desiderata, the literature has consolidated several major families of methods. 
\emph{Replay-based} methods mitigate forgetting by approximating access to past experience, either by retaining examples in a memory buffer with designed sample-selection rules~\cite{rebuffi2017icarl, rolnick2019experience}, generating pseudo-samples through generative replay, or storing intermediate representations through feature replay. 
\emph{Gradient-based} methods modify the optimization process itself~\cite{tanikanti2025first}, projecting updates onto directions that reduce interference with earlier tasks (\eg Gradient Episodic Memory (GEM)~\cite{lopezpaz2017gem}, Orthogonal Gradient Descent (OGD)~\cite{farajtabar2020orthogonal}), and C-Flat series~\cite{bian2024make, li2026faster}. 
\emph{Architecture-based} methods~~\cite{lu2025rethinking, lu2024revisiting} reduce interference by isolating or expanding capacity, allocating task-specific subspaces or modules through pruning, hard attention masks, or dynamic growth (\eg PackNet~\cite{mallya2018packnet}, Hard Attention to the Task (HAT)~\cite{serra2018hat}, and progressive networks~\cite{rusu2016progressive}).
\emph{Regularization-based} methods preserve prior knowledge by constraining updates~\cite{feng2022overcoming}, either through parameter-importance penalties based on Fisher information or related sensitivity measures (\eg Elastic Weight Consolidation (EWC)~\cite{Kirkpatrick_2017}, Synaptic Intelligence (SI)~\cite{zenke2017continual}, and Memory Aware Synapses (MAS)~\cite{mas}) or through functional regularization that distills the behavior of previous models into the current one (\eg Learning without Forgetting (LwF)~\cite{li2017learning}). 

Despite their distinct technical mechanisms, these method families largely operate within a common formulation that is often left implicit. Three assumptions are particularly important: (i) capability acquisition and retention primarily occur during a dedicated \emph{training} stage before deployment; (ii) accumulated capability is primarily carried by model \emph{parameters}; and (iii) updates are typically implemented through \emph{gradient}-based optimization over externally supplied or previously collected data. These assumptions are not intrinsic to the broader goal of continual learning; rather, they reflect the design choices of the setting in which the field originally developed. Together, they delineate the classical formulation and reveal the directions along which recent work has begun to extend it.

\subsection{From Three Assumptions to Three Questions}
\label{sec:three_axis_extension}

Each of the three assumptions is, in effect, a fixed answer to a question that any continual learning system must implicitly settle. Assumption (i) fixes the answer to \emph{when} learning happens: capability evolution is confined to a dedicated training stage. Assumption (ii) fixes the answer to \emph{where} capability resides: it is carried by model parameters. Assumption (iii) fixes the answer to \emph{how} the update is driven: by gradient-based optimization over externally supplied data. In the classical setting these answers were so uniform that the questions themselves were rarely asked. The developments reviewed in Section~\ref{sec:intro} make each answer a genuine variable: learning now happens at several stages of the model lifecycle, capability accumulates on carriers beyond parameters, and updates are driven by mechanisms beyond off-policy gradients. We therefore recast continual learning from sequential parameter learning to continual capability evolution, characterized through the When, Where, and How questions. The three dimensions are used as complementary analytical perspectives rather than as mutually exclusive categories. A method may be associated with more than one applicable label, and its characterization may change over time when capability moves between lifecycle stages or carriers. We therefore describe a method using a When–Where–How profile rather than treating it as occupying one immutable point in a strict Cartesian grid.

Within the taxonomy adopted in this survey, classical continual learning remains the reference setting of post-training in When, model parameters in Where, and off-policy gradient updates in How. The developments of the LLM era extend this reference setting: When spans the full model lifecycle, including continual pre-training and inference-time adaptation; Where extends from parameters to the harness; and How extends from off-policy gradient learning to on-policy learning and learning beyond gradients. Section 3 examines these three dimensions in turn.

 \section{The Three Dimensions of Continual Learning}
\label{sec:taxonomy}

\subsection{Overview}
\label{sec:taxonomy_overview}

We examine every continual learning method by asking three questions: when capability evolution takes place, where the acquired capability is accumulated, and how the update is driven. 

The three axes are intended as complementary perspectives rather than mutually exclusive categories. A continual learning method can be examined from all three at the same time, since its learning stage, capability carrier, and update mechanism jointly determine how continual capability evolution is realized. From this view, classical continual learning corresponds to a canonical setting, namely post-training updates to model parameters through off-policy gradient learning, while recent large-model and agent-based methods depart from this setting along one or more dimensions.

This perspective helps clarify that continual learning is no longer confined to the classical problem of preventing forgetting during sequential parameter training. Instead, it increasingly concerns continual capability evolution across the full lifecycle of large models and agent systems. Sections~\ref{sec:when} through \ref{sec:how} examine the three perspectives in turn (Figure~\ref{fig:method_tree}), and Section~\ref{sec:cross_axis_combinations} places existing methods back into the joint space of the three axes and analyzes how they combine and populate it.

%
\definecolor{bBlue}{RGB}{226,238,252}   \definecolor{lBlue}{RGB}{120,160,215}
\definecolor{bOrange}{RGB}{253,235,219}\definecolor{lOrange}{RGB}{228,148,86}
\definecolor{bGreen}{RGB}{228,245,230}  \definecolor{lGreen}{RGB}{128,188,140}
\definecolor{bGray}{RGB}{236,236,236}   \definecolor{lGray}{RGB}{150,150,150}

\providecommand{\mlist}[1]{%
  \parbox[t]{7.6cm}{\raggedright\sloppy\hbadness=10000
    \setlength{\emergencystretch}{3em}\hyphenpenalty=200
    \fontsize{7}{8.6}\selectfont #1\strut}}

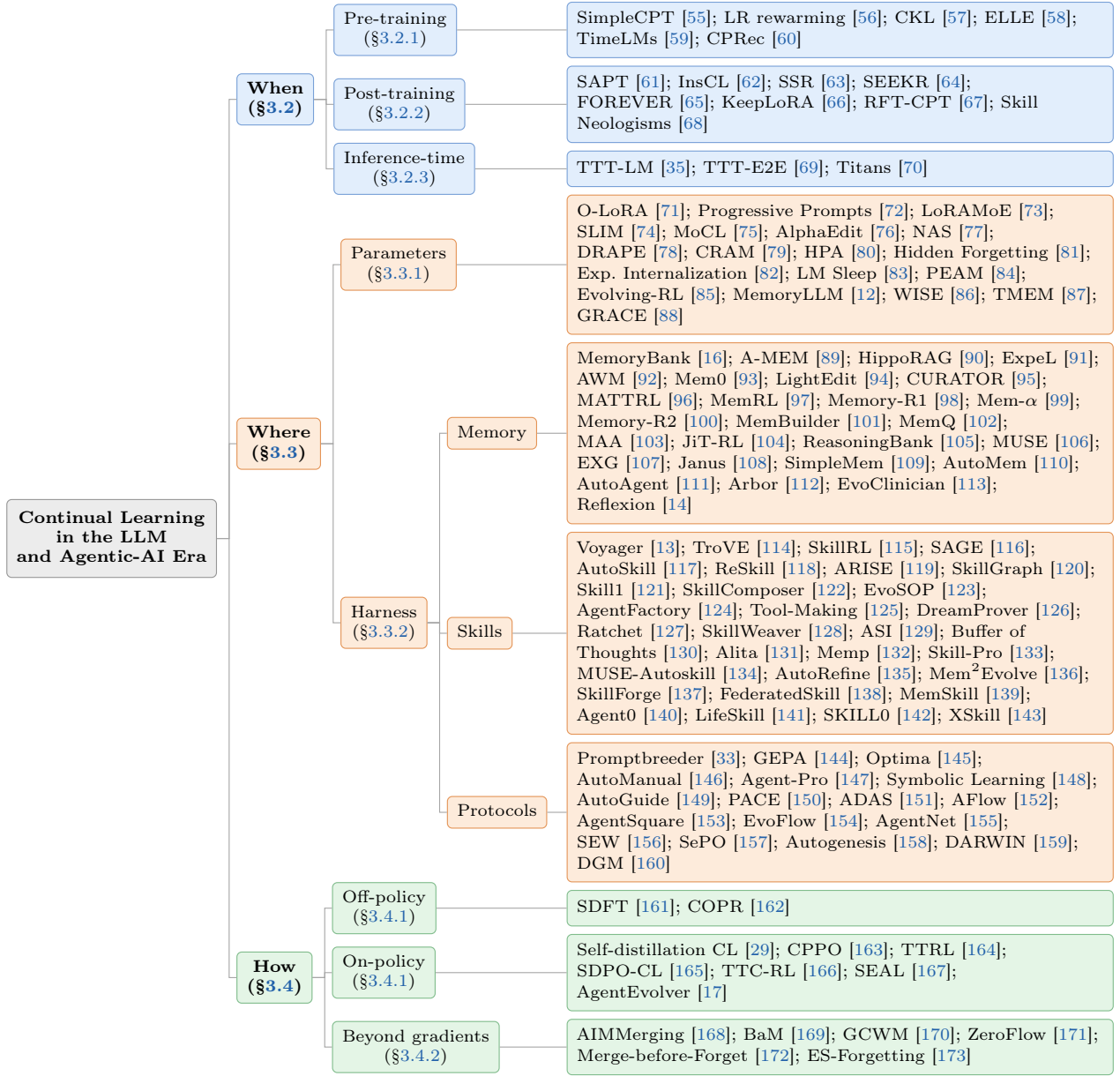
\begin{figure}[!t]
\centering
\resizebox{\textwidth}{!}{%
\begin{forest}
  forked edges,
  for tree={
    grow'=east,
    thin,
    rounded corners=2pt,
    draw,
    anchor=west,
    child anchor=west,
    parent anchor=east,
    inner xsep=4pt,
    inner ysep=2.5pt,
    s sep=3.1pt,
    l sep=8pt,
    edge={thin,gray!70},
    font=\scriptsize,
    align=center,
  },
  delay={where n children=0{
    tier=methodlist,
    inner ysep=3pt,
  }{}},
  where level=0{
    fill=bGray,
    draw=lGray,
    font=\scriptsize\bfseries,
    inner ysep=4pt
  }{},
  where level=1{
    font=\scriptsize\bfseries
  }{},
  [{Continual Learning\\in the LLM\\and Agentic-AI Era}
    [{When\\(\S\ref{sec:when})}, fill=bBlue, draw=lBlue
      [{Pre-training\\(\S\ref{sec:pre_training})}, fill=bBlue, draw=lBlue
        [{\mlist{
          SimpleCPT~\cite{ibrahim2024simple};
          LR rewarming~\cite{gupta2023continual};
          CKL~\cite{jang2022towards};
          ELLE~\cite{qin2022elle};
          TimeLMs~\cite{loureiro2022timelms};
          CPRec~\cite{CPRec}
        }}, fill=bBlue, draw=lBlue]
      ]
      [{Post-training\\(\S\ref{sec:post_training})}, fill=bBlue, draw=lBlue
        [{\mlist{
          SAPT~\cite{zhao2024sapt};
          InsCL~\cite{wang2024inscl};
          SSR~\cite{huang2024ssr};
          SEEKR~\cite{he2024seekr};
          FOREVER~\cite{feng2026forever};
          KeepLoRA~\cite{luo2026keeplora};
          RFT-CPT~\cite{zhang2025reinforcement};
          Skill Neologisms~\cite{berthon2026skillneologisms}
        }}, fill=bBlue, draw=lBlue]
      ]
      [{Inference-time\\(\S\ref{sec:inference_time})}, fill=bBlue, draw=lBlue
        [{\mlist{
          TTT-LM~\cite{sun2024learning};
          TTT-E2E~\cite{tandon2025e2ettt};
          Titans~\cite{behrouz2025titans}
        }}, fill=bBlue, draw=lBlue]
      ]
    ]
    [{Where\\(\S\ref{sec:where})}, fill=bOrange, draw=lOrange
      [{Parameters\\(\S\ref{sec:parametric_carrier})}, fill=bOrange, draw=lOrange
        [{\mlist{
          O-LoRA~\cite{wang2023orthogonal};
          Progressive Prompts~\cite{razdaibiedina2023progressive};
          LoRAMoE~\cite{dou2024loramoe};
          SLIM~\cite{han2025slim};
          MoCL~\cite{wang2024rehearsalfree};
          AlphaEdit~\cite{fang2025alphaedit};
          NAS~\cite{liu2026norm};
          DRAPE~\cite{drape2026};
          CRAM~\cite{cram2026};
          HPA~\cite{hpa2025};
          Hidden Forgetting~\cite{rclhidden2026};
          Exp.\ Internalization~\cite{expinternalize2026};
          LM Sleep~\cite{lmsleep2026};
          PEAM~\cite{peam2026};
          Evolving-RL~\cite{evolvingrl2026};
          MemoryLLM~\cite{wang2024memoryllm};
          WISE~\cite{wang2024wise};
          TMEM~\cite{ren2026parametric};
          GRACE~\cite{hartvigsen2023grace}
        }}, fill=bOrange, draw=lOrange]
      ]
      [{Harness\\(\S\ref{sec:harness_layer})}, fill=bOrange, draw=lOrange
        [{Memory}, fill=bOrange, draw=lOrange
          [{\mlist{
            MemoryBank~\cite{zhong2024memorybank};
            A-MEM~\cite{xu2025amem};
            HippoRAG~\cite{gutierrez2024hipporag};
            ExpeL~\cite{zhao2024expel};
            AWM~\cite{wang2025awm};
            Mem0~\cite{chhikara2025mem0};
            LightEdit~\cite{jung2026lightedit};
            CURATOR~\cite{wu2026curator};
            MATTRL~\cite{hu2026mattrl};
            MemRL~\cite{memrl2026};
            Memory-R1~\cite{memoryrone2025};
            Mem-$\alpha$~\cite{memalpha2025};
            Memory-R2~\cite{memoryrtwo2026};
            MemBuilder~\cite{membuilder2026};
            MemQ~\cite{memq2026};
            MAA~\cite{maaccum2026};
            JiT-RL~\cite{jitrl2026};
            ReasoningBank~\cite{reasoningbank2025};
            MUSE~\cite{musejob2025};
            EXG~\cite{exg2026};
            Janus~\cite{janusprologue2026};
            SimpleMem~\cite{simplemem2026};
            AutoMem~\cite{automem2026};
            AutoAgent~\cite{autoagentmem2026};
            Arbor~\cite{arbor2026};
            EvoClinician~\cite{evoclinician2026};
            Reflexion~\cite{shinn2023reflexion}
          }}, fill=bOrange, draw=lOrange]
        ]
        [{Skills}, fill=bOrange, draw=lOrange
          [{\mlist{
            Voyager~\cite{wang2023voyager};
            TroVE~\cite{wang2024trove};
            SkillRL~\cite{xia2026skillrl};
            SAGE~\cite{wang2025sage};
            AutoSkill~\cite{yang2026autoskill};
            ReSkill~\cite{reskill2026};
            ARISE~\cite{arise2026};
            SkillGraph~\cite{skillgraph2026};
            Skill1~\cite{skillone2026};
            SkillComposer~\cite{skillcomposer2026};
            EvoSOP~\cite{evosop2026};
            AgentFactory~\cite{agentfactory2026};
            Tool-Making~\cite{toolmaking2026};
            DreamProver~\cite{dreamprover2026};
            Ratchet~\cite{ratchet2026};
            SkillWeaver~\cite{skillweaver2025};
            ASI~\cite{asiinduction2025};
            Buffer of Thoughts~\cite{bufferofthoughts2024};
            Alita~\cite{alita2025};
            Memp~\cite{memp2025};
            Skill-Pro~\cite{procmem2026};
            MUSE-Autoskill~\cite{museautoskill2026};
            AutoRefine~\cite{autorefine2026};
            Mem$^2$Evolve~\cite{memtwoevolve2026};
            SkillForge~\cite{skillforge2026};
            FederatedSkill~\cite{federatedskill2026};
            MemSkill~\cite{memskill2026};
            Agent0~\cite{agent02025};
            LifeSkill~\cite{mao2026lifeskill};
            SKILL0~\cite{lu2026skill0};
            XSkill~\cite{xskill2026}
          }}, fill=bOrange, draw=lOrange]
        ]
        [{Protocols}, fill=bOrange, draw=lOrange
          [{\mlist{
            Promptbreeder~\cite{fernando2023promptbreeder};
            GEPA~\cite{agrawal2025gepa};
            Optima~\cite{optima2024};
            AutoManual~\cite{automanual2024};
            Agent-Pro~\cite{agentpro2024};
            Symbolic Learning~\cite{agentsymbolic2024};
            AutoGuide~\cite{autoguide2024};
            PACE~\cite{pace2026};
            ADAS~\cite{adas2024};
            AFlow~\cite{aflow2024};
            AgentSquare~\cite{agentsquare2024};
            EvoFlow~\cite{evoflow2025};
            AgentNet~\cite{agentnet2025};
            SEW~\cite{sew2025};
            SePO~\cite{sepo2026};
            Autogenesis~\cite{autogenesis2026};
            DARWIN~\cite{darwin2026};
            DGM~\cite{dgm2025}
          }}, fill=bOrange, draw=lOrange]
        ]
      ]
    ]
    [{How\\(\S\ref{sec:how})}, fill=bGreen, draw=lGreen
      [{Off-policy\\(\S\ref{sec:update_regime})}, fill=bGreen, draw=lGreen
        [{\mlist{
          SDFT~\cite{yang2024sdft};
          COPR~\cite{zhang2025copr}
        }}, fill=bGreen, draw=lGreen]
      ]
      [{On-policy\\(\S\ref{sec:update_regime})}, fill=bGreen, draw=lGreen
        [{\mlist{
          Self-distillation CL~\cite{shenfeld2026sdft};
          CPPO~\cite{zhang2024cppo};
          TTRL~\cite{ttrl2025};
          SDPO-CL~\cite{sdpocl2026};
          TTC-RL~\cite{hubotter2025learning};
          SEAL~\cite{zweiger2025seal};
          AgentEvolver~\cite{zhai2025agentevolver}
        }}, fill=bGreen, draw=lGreen]
      ]
      [{Beyond gradients\\(\S\ref{sec:beyond_gradients})}, fill=bGreen, draw=lGreen
        [{\mlist{
          AIMMerging~\cite{feng2025aimmerging};
          BaM~\cite{alexandrov2024bam};
          GCWM~\cite{wang2026geometry};
          ZeroFlow~\cite{zeroflow2025};
          Merge-before-Forget~\cite{mergebeforeforget2025};
          ES-Forgetting~\cite{esforgetting2026}
        }}, fill=bGreen, draw=lGreen]
      ]
    ]
  ]
\end{forest}%
}
\caption{Representative LLM-era and agentic-AI-era continual-learning methods
organized along the three dimensions. Methods that span several dimensions are
placed once, under their most salient carrier or mechanism.}
\label{fig:method_tree}
\end{figure}


\begin{figure}[!t]
  \centering
  \includegraphics[width=0.9\linewidth]{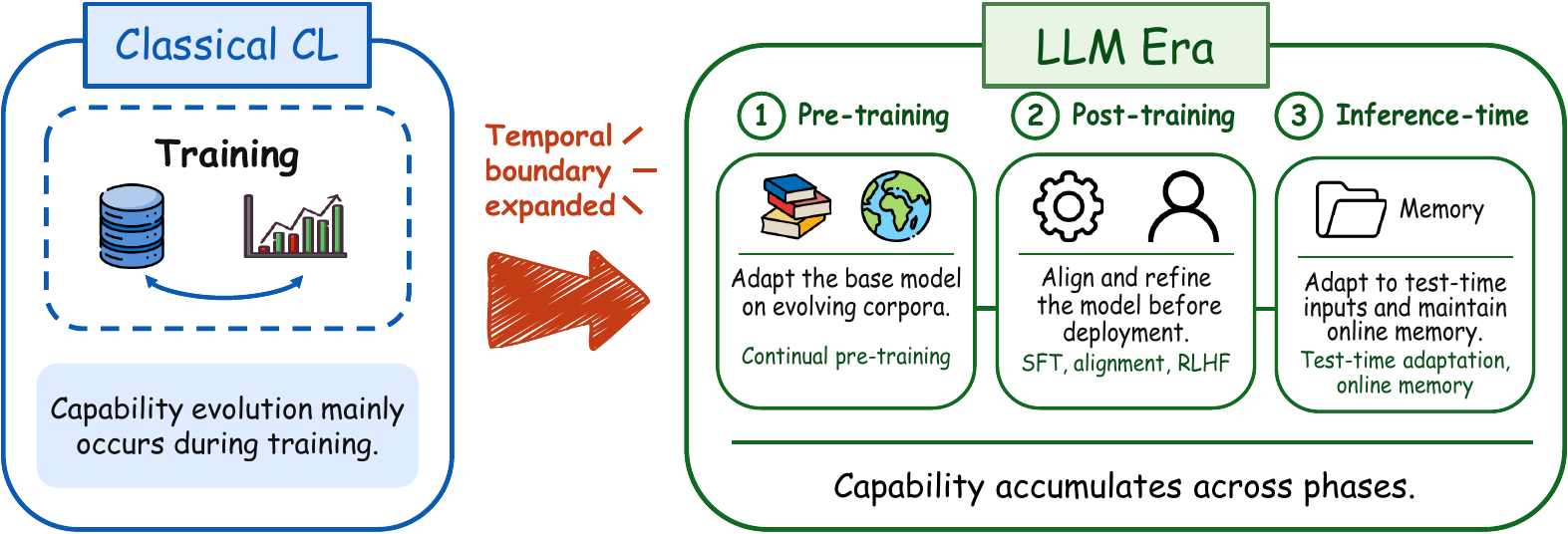}
  \vspace{-0.2cm}
  \caption{The \When dimension. Classical continual learning confines capability evolution to a single training stage (left). In the LLM era, it spans pre-training, post-training, and inference time, so that capability accumulates across the phases of the model lifecycle (right).}
  \label{fig:when_axis}
\end{figure}

\subsection{When: Capability Evolution across the Model Lifecycle}
\label{sec:when}

The \When axis asks at which stage of the model lifecycle continual capability evolution is realized. In the era of large models, this question becomes especially important because learning is no longer confined to a single sequential training process (Figure~\ref{fig:when_axis}). Instead, capability evolution may occur during continued pre-training over evolving corpora, during post-training through instruction tuning, preference optimization, or reinforcement learning (RL), and after deployment through inference-time adaptation to incoming inputs and environmental feedback.

This temporal shift changes not only the location of continual learning, but also the form of the problem itself. Different lifecycle stages expose different update targets, feedback signals, and stability requirements. Continued pre-training emphasizes knowledge renewal under changing data distributions. Post-training emphasizes capability alignment, task adaptation, and forgetting control across successive optimization stages. Inference-time adaptation emphasizes rapid adjustment under test-time distribution shift, often with limited writable state or lightweight parameter updates. The \When axis therefore repositions continual learning as a lifecycle-level problem of how large models and agent systems keep evolving over time.

\subsubsection{Pre-Training}
\label{sec:pre_training}

Continual pre-training turns pre-training itself from a one-shot procedure into a multi-stage process over evolving corpora, so that a model already trained on a base distribution can be further adapted to new domains, languages, or temporal slices without restarting from scratch. The direction has roots in domain- and task-adaptive pre-training, which showed that a further phase of pre-training on domain or task corpora consistently improves downstream performance~\cite{gururangan2020dont}, and in continual knowledge learning, which formulates keeping a language model's world knowledge current as renewing outdated facts while retaining time-invariant ones~\cite{jang2022towards}. Ibrahim~\etal~\cite{ibrahim2024simple} establish a simple yet scalable recipe that combines learning-rate rewarming with a small replay buffer of base-distribution tokens, allowing 405M and 10B models to absorb new corpora while keeping degradation on the original distribution close to negligible. Y{\i}ld{\i}z~\etal~\cite{yildiz2024investigating} complement this with a systematic study of how the magnitude and ordering of domain shifts modulate forgetting and forward transfer at LLM scale, building on earlier empirical evidence on warm-up schedules from~\cite{gupta2023continual}. Relative to the classical setting in which task boundaries are explicit and supervised losses are reused, continual pre-training operates over unsupervised next-token objectives at scales where individual task labels are absent, shifting the unit of incremental adaptation from a labeled task to a corpus stream.

\subsubsection{Post-Training}
\label{sec:post_training}

After pre-training, post-training has evolved from a single instruction-tuning step into a longitudinal pipeline: supervised fine-tuning is typically followed by one or more preference- or reward-alignment stages (\eg RLHF, RLVR), and deployed models continue to be revised in successive versions. Each round of updating risks eroding what earlier rounds established, which makes this stage a continual-learning problem in its own right.

Recent work characterizes this problem empirically. Luo~\etal~\cite{Luo2023AnES} quantify catastrophic forgetting during continual fine-tuning of LLMs, showing that domain knowledge, reasoning, and reading comprehension all degrade as instruction tuning proceeds, with the effect intensifying with model scale in the 1B--7B range. TRACE~\cite{trace} consolidates this observation into a benchmark for aligned LLMs, demonstrating that sequential training on new tasks erodes not only general ability but also instruction following and safety alignment. Zheng~\etal~\cite{zheng2025spuriousforgettingcontinuallearning} refine the diagnosis by showing that part of the measured degradation is \emph{spurious} forgetting, a recoverable loss of task alignment rather than a loss of underlying knowledge, and that freezing bottom layers largely prevents it. The alignment stages contribute their own failure mode: preference optimization imposes an alignment tax on upstream capabilities, which InstructGPT already mitigated by mixing pre-training gradients into RLHF~\cite{ouyang2022training}, and the choice of update mechanism strongly modulates how much a given stage forgets~\cite{rlsRazor2025}. A related effect appears at the interface between post-training paradigms: ReCALL~\cite{yang2026recall} shows that adapting a generative MLLM into a discriminative retriever can degrade its native fine-grained reasoning, and uses a self-guided diagnose--generate--refine loop to mine corrective triplets and realign the retriever through continued training. Although ReCALL is not formulated as a sequential continual-learning benchmark, it illustrates that stability problems can arise between capability regimes as well as between explicitly delineated tasks. Together, these findings establish the post-training pipeline itself, rather than any single fine-tuning step, as the unit at which forgetting must be measured and controlled~\cite{liu2026continual}.

\subsubsection{Inference-Time}
\label{sec:inference_time}

The \emph{inference-time} position on the \When axis refers to capability evolution that occurs after deployment. The key distinction is persistence. General inference-time computation, such as search, sampling, self-consistency, or other test-time scaling strategies, can improve the answer to a single query, but it usually leaves the system unchanged once the query is completed. Inference-time continual learning, in contrast, writes information obtained during inference back into the system, through writable states or parameter updates, so that earlier inputs, feedback signals, or self-generated supervision can influence later behavior.

This setting is especially relevant for large models and agent systems. During deployment, a model may face long input streams, shifting user requirements, evolving task distributions, and feedback signals that are unavailable during training. Treating inference as a purely read-only process limits the system to static deployment. Inference-time continual learning instead turns deployment into an adaptive process, where the model or its surrounding system can accumulate information from the test stream and adjust its behavior over time.

Recent methods instantiate this idea in different ways. TTT-LM~\cite{sun2024learning} updates trainable hidden-state modules within a test sequence, allowing contextual information from earlier tokens to be compressed into fast weights. TTT-E2E~\cite{tandon2025e2ettt} extends this direction to long-context language modeling by updating model components through next-token prediction at inference time, thereby using parameter updates to absorb information from the growing context. TTRL~\cite{ttrl2025} further introduces reward-driven test-time learning, deriving supervision from the model's own sampled outputs on unlabeled test inputs and using it to improve performance on the test distribution.

From the perspective of the \When axis, these methods mark a shift from static deployment to post-deployment capability evolution. They expand continual learning beyond pre-deployment optimization and show that adaptation can also occur during the actual use of a model, as long as inference-time signals are persistently accumulated and affect subsequent predictions.


\subsection{Where: Capability Carriers from Parameters to the Harness}
\label{sec:where}

\begin{figure}[!t]
  \centering
  \includegraphics[width=0.9\linewidth]{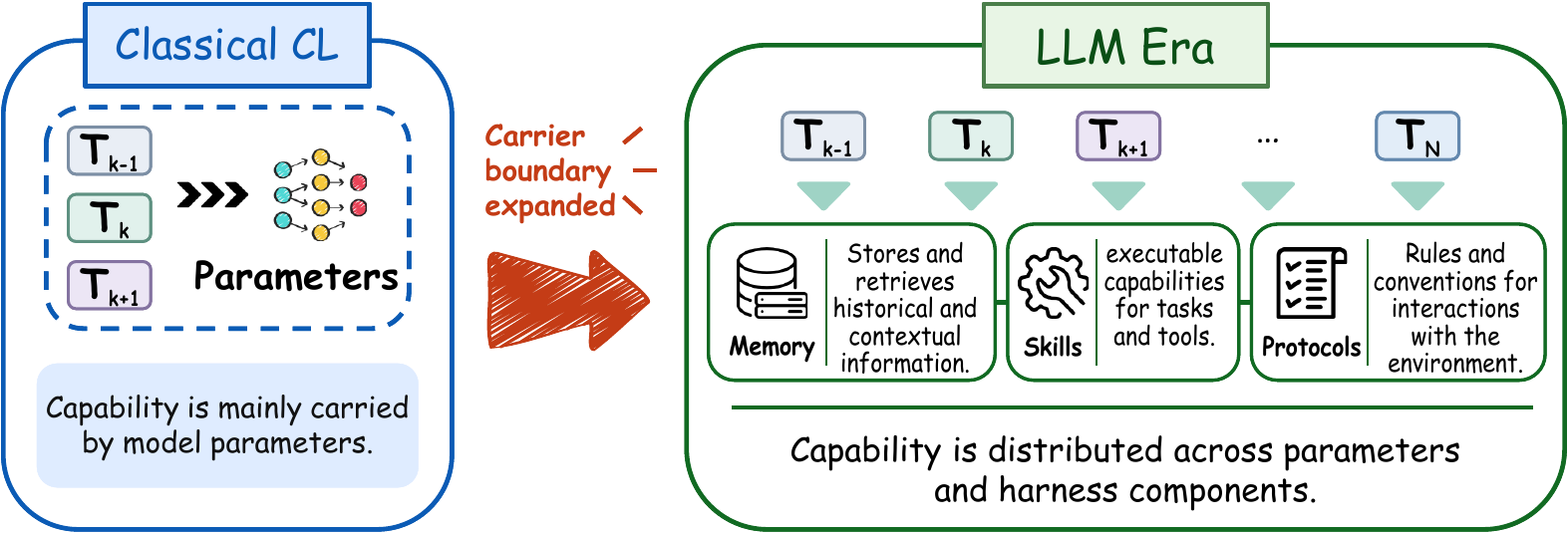}
  \vspace{-0.2cm}
  \caption{The \Where dimension. Classical continual learning carries capability almost exclusively in model parameters (left); in the LLM era, capability is distributed across parameters and the harness components, namely memory, skills, and protocols (right).}
  \label{fig:where_axis}
\end{figure}

The \Where axis asks where acquired knowledge and capabilities are stored and updated. Model parameters remain the canonical carrier of continual learning, although modern systems may update them through full fine-tuning, parameter-efficient modules, or other structured forms. The more fundamental shift in the LLM era, however, is that capability accumulation increasingly extends beyond the parameter space (Figure~\ref{fig:where_axis}). Agent systems externalize part of their evolving capability into the harness layer surrounding the core model, including memory, skills, and protocols. These external carriers are explicit, editable, retrievable, and callable, allowing continual learning to operate not only through weight updates, but also through persistent non-parametric objects that can be accumulated and revised over time.

\subsubsection{Parametric Carrier}
\label{sec:parametric_carrier}

Parameters remain the canonical carrier of continual learning: new capabilities are consolidated by modifying trainable weights, and forgetting is often studied as interference within the shared parameter space. In the LLM era, however, the main change lies in the granularity of parametric updates. Full-parameter fine-tuning is often costly and prone to interference at scale, so many continual adaptation methods restrict updates to parameter-efficient or task-specific modules. This follows the motivation of reducing cross-task interference under LLM-scale constraints, while avoiding holistic modification of the entire model.

The low-rank adaptation (LoRA)~\cite{hu2022lora} family exemplifies this direction. O-LoRA~\cite{wang2023orthogonal} constrains the LoRA directions of different tasks to be mutually orthogonal during continual fine-tuning, thereby reducing overwriting between task-specific subspaces. Similar ideas also appear across the broader parameter-efficient fine-tuning (PEFT) family. Progressive Prompts~\cite{razdaibiedina2023progressive} learns a separate soft prompt for each task and concatenates it with previously learned prompts, confining adaptation to separable prompt parameters. Expert-based approaches such as LoRAMoE~\cite{dou2024loramoe} route different inputs or tasks to different LoRA experts, while SLIM~\cite{han2025slim} combines LoRA experts with an identity pathway to support new-task learning with reduced forgetting. MemoryLLM~\cite{wang2024memoryllm} extends the parametric carrier to continual knowledge storage by augmenting a fixed backbone with a learnable memory pool of fixed capacity. Incoming information updates the pool's latent memory tokens during inference. These tokens participate directly in internal computation and constitute the method's evolving parametric state.

These methods refine the parametric carrier from a monolithic parameter space into a set of modular, locally updated components. Nevertheless, the acquired capability is still stored in trainable parameters inside or attached to the model. The more substantial expansion of the \Where axis begins when capability accumulation moves beyond parametric objects to the external harness layer.

\subsubsection{The Harness Layer}
\label{sec:harness_layer}

With the rapid development of LLM agent systems, the carrier of knowledge and capability is no longer confined to model parameters but extends further to the harness layer surrounding the core model~\cite{wang2026harness}, namely a collection of objects external to the parameters that are explicitly constructed and edited by the system designer. Following~\cite{externalization2026}, the harness divides, by the nature of what it carries, into three classes: \emph{memory} is information storage that is readable and writable across inference calls (such as dialogue history, long-term memory banks, and retrieved document fragments); \emph{skills} are executable units that the model actively invokes (such as external tools, function modules, and callable sub-agents); and \emph{protocols} are rules and formats that govern how the model interacts with its environment (such as system prompts, rule files, and message-format schemas). All three classes can be read, extended, or rearranged without updating the backbone weights, supporting continual learning outside the parameter tensor. In the LLM era, each has given rise to a relatively mature methodological lineage, covering hierarchical designs of memory systems, externalized extension of tool repertoires and skill libraries, and the evolution of prompt-, rule-, and message-format protocols, respectively.

\noindent\textbf{Memory.}\quad As the harness component that carries historical and contextual information, memory provides LLM systems with a pathway for capability accumulation that does not depend on gradient updates: the model writes facts, experience, and intermediate states into external storage and retrieves them in subsequent inference, thereby continually expanding the body of knowledge it can access and preserving historical context without modifying backbone weights. Recent work pushes memory as a continual-learning carrier along several distinct design dimensions. \textit{MemoryBank}~\cite{zhong2024memorybank} consolidates long-term memory for dialogue agents through an Ebbinghaus-style forgetting curve that strengthens important information across interactions while gradually fading redundant content, making the stability--plasticity trade-off of continual learning explicit at the memory-consolidation layer. \textit{A-MEM}~\cite{xu2025amem} takes atomic propositions as the smallest unit of reading, writing, and forgetting, reducing the granularity of memory updates to the fact level and avoiding the collateral forgetting that paragraph-level overwrites induce. \textit{MemRL}~\cite{memrl2026} extends the memory-update mechanism from rule-based retrieval to reward-driven RL self-evolution. Collectively these methods turn memory from a passive context container into a writable, structurable, self-managed capability carrier in which capability evolution proceeds outside the parameter tensor and accumulates across tasks.

\noindent\textbf{Skills.}\quad As the harness component that carries procedural capability, skills encapsulate the model's reusable execution flows as externally callable units. By adding, refining, and composing skill units, an agent extends its action repertoire without updating backbone weights, and continual learning proceeds as skill accumulation at the execution layer. Representative methods advance the skill library as a continual-learning carrier along different design dimensions. \textit{Voyager}~\cite{wang2023voyager} lets an LLM agent write code as reusable skills while interacting with its environment and store them in a persistent skill library, so that the agent continually masters new tasks through progressive expansion of the library while the discreteness of skills keeps new and old tasks from interfering. \textit{SkillRL}~\cite{xia2026skillrl} jointly optimizes the skill library and the policy via reinforcement learning, extending skill updates from rule-based curation to reward-driven iterative refinement so that the growth of the library is also guided by task utility. \textit{SAGE}~\cite{wang2025sage} couples the skill library with reinforcement learning to build a self-improving agent that acquires, stores, and reuses skills under reward feedback, so that the library grows with the agent's own experience. Beyond these methods that take the skill library as the update target, \textit{SKILL0}~\cite{lu2026skill0} further illustrates a reverse migration between skills and the parametric carrier by first accumulating skills on the harness via RL and then internalizing them into model parameters via curriculum (discussed in Section~\ref{sec:cross_axis_combinations}). Together these methods turn the skill library from a static toolkit into an extensible, optimizable carrier that can also migrate to parameters, enabling the agent's capability repertoire to accumulate continuously across tasks.

\noindent\textbf{Protocols.}\quad As the harness component that carries interaction structure, protocols externalize the behavioral rules and communication conventions between the model and its environment as editable artifacts. By revising and self-evolving these rules, an agent continually adjusts its behavioral framework without updating backbone weights, allowing continual learning to proceed in the form of protocol evolution at the interaction layer. \textit{Promptbreeder}~\cite{fernando2023promptbreeder} uses an LLM to evolve prompts themselves through self-referential mutation, making the system prompt a carrier that task feedback continually rewrites; it is the most explicit representative of continual learning via protocol evolution. Rule files follow the same externalization approach: engineering practices such as Cline, Aider, and Cursor externalize an agent's behavioral rules as files that developers iteratively revise as project requirements evolve, treating the interaction structure as a continually editable carrier. Compared with the denser lineages already formed for memory and skills, the research landscape of protocols as a continual-learning carrier remains at an early stage, but it is a necessary conceptual anchor among the three classes of externalization that the harness comprises.


\subsection{How: Update Mechanisms from Off-Policy Gradients to Gradient-Free Learning}
\label{sec:how}

\begin{figure}[!t]
  \centering
   \includegraphics[width=0.9\linewidth]{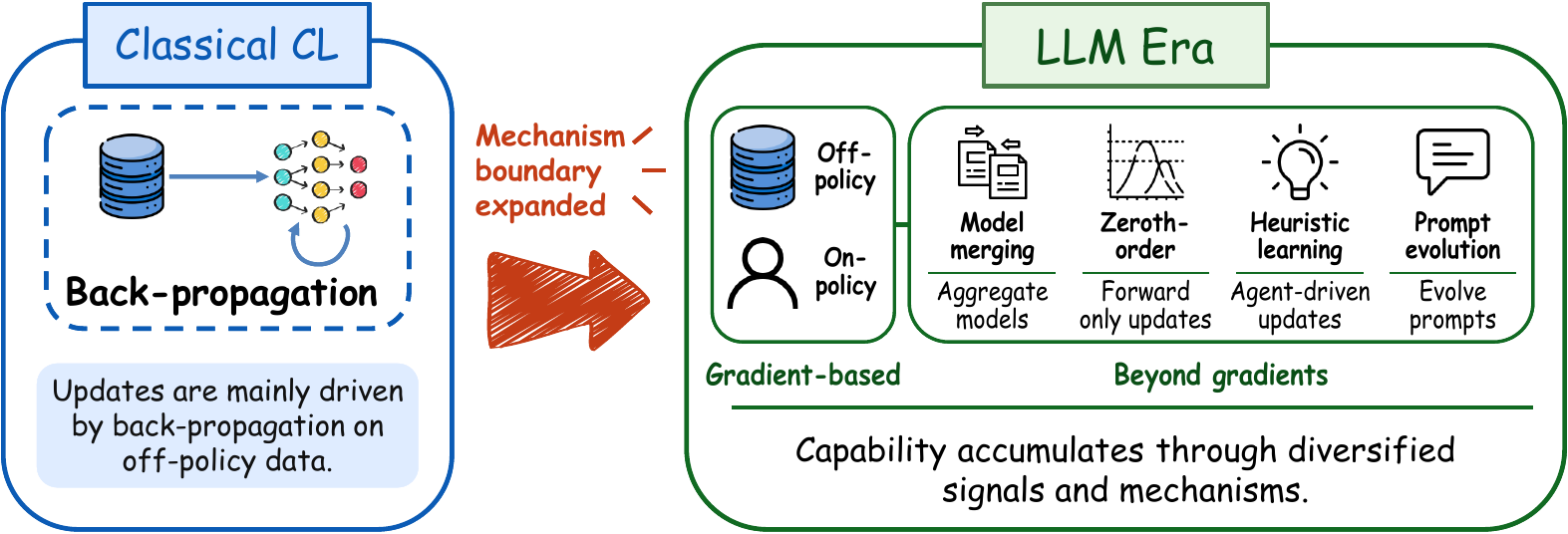}
  \vspace{-0.2cm}
  \caption{The \How dimension. Classical continual learning relies on backpropagation over off-policy data (left); the LLM era adds on-policy learning within the gradient-based regime, as well as mechanisms for learning beyond gradients, such as model merging, zeroth-order optimization, heuristic learning, and prompt evolution (right).}
  \label{fig:how_axis}
\end{figure}

The LLM era extends the \How axis in two directions (Figure~\ref{fig:how_axis}). First, within gradient-based learning, it broadens the data source from off-policy data decoupled from the current policy (historical replay, static corpora, distillation targets) to on-policy rollouts generated by the policy that is currently being updated under reward signals (Section~\ref{sec:update_regime}). Second, it moves beyond backpropagated gradients altogether, accumulating capability through mechanisms that never compute an analytic gradient of a loss (Section~\ref{sec:beyond_gradients}). What unifies the continual-learning relevance of these regimes is how each controls drift away from prior capability: the central question along the \How axis is not merely how an update is computed, but how strongly it perturbs what the model already knows.

\subsubsection{Update Regime: Off-Policy vs.\ On-Policy}
\label{sec:update_regime}

The off-policy methods in classical continual learning are dominated by replay over historical samples and distillation against the outputs of an older model, both relying on externally supplied signal sources. The continual-learning question for off-policy updates is how to fine-tune on new data without drifting away from previously acquired behavior. Self-distillation offers a direct LLM-era answer: SDFT~\cite{yang2024sdft} prompts the model to rewrite the task data into responses that match its own distribution and then fine-tunes on this self-generated, distribution-aligned data, thereby bridging the distribution gap between the task data and the base model and empirically mitigating catastrophic forgetting of general capabilities and safety alignment. Off-policy learning thus controls drift by keeping the training distribution close to the model's own.

The genuinely new mechanism that the LLM era brings to the \How axis is on-policy learning, where training data is generated by the very policy being updated, conditioned on reward signals; the reward-driven alignment paradigms of post-training (\eg RLHF,  RLVR) are its principal instances. Its significance for continual learning is structural rather than incidental: Shenfeld~\etal~\cite{rlsRazor2025} show that the degree of forgetting depends on the Kullback--Leibler (KL) divergence between the fine-tuned and the reference policy on the new task, and that on-policy RL implicitly prefers, among the many solutions to a new task, the one with smallest KL, so its drift away from previously acquired capabilities stays small---whereas off-policy supervised fine-tuning can converge to distributions arbitrarily far from the reference model. On-policy learning is therefore not merely an additional mechanism but one structurally aligned with the central concern of continual learning, avoiding forgetting; at inference time the same principle is carried into deployment by TTRL~\cite{ttrl2025}, which drives test-time updates with an on-policy reward derived from the model's own majority vote.

On-policy updates can also be differentiated at the sample level. SRPO~\cite{li2026unifying} routes correct current-policy rollouts to group-relative reward optimization and failed rollouts to targeted self-distillation, while entropy-aware weighting suppresses unreliable self-teacher signals. In this way, it unifies the reward alignment of GRPO with the dense corrective signal of self-distillation and avoids the latter's collapse during prolonged optimization. This is evidence about optimization stability rather than cross-task retention, but the routing principle is relevant to continual learning because it exposes a finer mechanism for controlling which update signal perturbs the policy on each new sample.

\subsubsection{Learning Beyond Gradients}
\label{sec:beyond_gradients}

Beyond the gradient-based regimes lies a family of mechanisms that accumulate capability without ever backpropagating an analytic gradient of a loss. The most established member is model merging, which composes capability directly in weight space among already-trained tensors. The pioneering Task Arithmetic~\cite{ilharco2022editing} defines task vectors as the differences between task-specific fine-tuned weights and the base weights and obtains a multi-task model by arithmetic combination of these vectors; TIES-Merging~\cite{yadav2023tiesmerging} prunes small-magnitude changes and aligns sign-consistent directions to reduce conflicts, Evolutionary Model Merge~\cite{akiba2024evolutionary} searches the parameter and data-flow spaces for an optimal recipe, and further methods such as DARE~\cite{yu2024dare} and Twin-Merging~\cite{lu2024twinmerging} refine coefficient control, aggregation granularity, and conflict mitigation. From a continual-learning standpoint, model merging accumulates multi-task capability with zero access to historical training data, directly addressing the no-historical-data constraint of classical continual learning; recent work such as AIMMerging~\cite{feng2025aimmerging} connects it explicitly to the continual-learning problem for language models, and a dedicated survey~\cite{yang2024modelmerging} reviews the area, which this survey reads as the canonical gradient-free mechanism along the \How axis.

A second, near-gradient member is forward-only or zeroth-order optimization, which updates weights by estimating a descent direction from forward passes alone, without backpropagation. MeZO~\cite{malladi2023mezo} established that a memory-efficient zeroth-order estimator can fine-tune large models with only forward evaluations, and ZeroFlow~\cite{zeroflow2025} showed that such forward-pass optimization is an effective continual learner: across forgetting benchmarks the implicit smoothing of zeroth-order updates matches or exceeds first-order fine-tuning at mitigating catastrophic forgetting---an intrinsic stability property paralleling the small-KL behavior of on-policy RL. Beyond weight space, the same gradient-free principle operates on the harness, where memory writes, skill accumulation, and prompt evolution update capability without backpropagation---a paradigm recently articulated as \emph{heuristic learning}, in which an agent revises an external system from feedback rather than gradient signals~\cite{weng2026learning}. Because for those methods the defining departure is the externalized carrier rather than the mechanism, they are detailed under the \Where axis (Section~\ref{sec:where}); here it suffices to note that learning beyond gradients spans a spectrum, from zeroth-order optimization that still estimates a gradient, through gradient-free weight composition, to harness edits that leave the parameter space entirely.

\subsection{Cross-Dimensional Method Profiles}
\label{sec:cross_axis_combinations}

Sections~\ref{sec:when}--\ref{sec:how} examine the literature one dimension at a time. Figure~\ref{fig:method_tree} places each representative method once, under the dimension that best captures its most salient carrier or mechanism. This primary placement does not encode the method's complete profile; Appendix~\ref{app:method-profiles} records the corresponding cross-dimensional profiles in Table~\ref{tab:method-profiles}.

Several methods illustrate why the three dimensions should be read together. Reflexion~\cite{shinn2023reflexion} combines inference-time evolution, a harness-based memory carrier, and learning beyond gradients through reflective text generation and direct memory writing. TTRL~\cite{ttrl2025} combines inference-time parameter adaptation with on-policy learning. AgentEvolver~\cite{zhai2025agentevolver} spans parameters and the harness: on-policy reinforcement learning updates the policy parameters, while selected trajectories are accumulated in experience memory. These methods are therefore described by combinations of labels rather than by their position in only one branch of Figure~\ref{fig:method_tree}.

A method's profile may also change over time. SKILL0~\cite{lu2026skill0} first accumulates reusable capability in a harness-based skill library and subsequently internalizes selected skills into model parameters through a curriculum. It is therefore represented as a Harness (Skills)-to-Parameters trajectory along the Where dimension rather than as a method with one fixed carrier.

Figures~\ref{fig:taxonomy_overview} contains selected representative methods and do not result from an exhaustive literature-enumeration protocol. Consequently, the number of methods displayed in a region should not be interpreted as a statistical estimate of research density, nor should an unoccupied region be treated as evidence that no relevant work exists. The figures instead illustrate recurring combinations and motivate comparatively underexplored questions for future research.

\section{Future Outlook and Discussion}
\label{sec:future_outlook}
This survey has reviewed how continual learning in the era of large models has spread well beyond its classical setting: capability is now updated at different points in the model lifecycle, carried by parameters as well as by external memory, skills, and protocols, and driven by a widening range of update mechanisms. Building on that overall picture, rather than on any single method, we step back and discuss several open trends together with the questions they raise for future work. Our central observation is that the field is shifting from an isolated algorithmic problem, namely the suppression of catastrophic forgetting in a single model, toward a broader question of how capability is organized across a whole system: where it should live, when it should be updated, and how it should be combined, retained, and released. We organize the discussion around five themes: the structural ceilings that limit how far capability can be pushed into ever longer context (Section~\ref{sec:outlook_carriers}); whether the hand-tuned engineering that currently manages these carriers already amounts to a learning mechanism (Section~\ref{sec:outlook_engineering}); how a model and its harness might instead evolve in a coordinated way (Section~\ref{sec:outlook_coevolution}); why long-horizon agents, rather than static benchmarks, form the real testbed for such directions (Section~\ref{sec:outlook_longhorizon}); and what all of this suggests for the path toward artificial general intelligence (AGI) and for research priorities (Section~\ref{sec:outlook_agi}).


\subsection{From Longer Context to Persistent Capability}
\label{sec:outlook_carriers}

A defining change of the LLM era is that the carrier of capability is no longer confined to the parameter tensor but migrates outward, from parameters to external memory and then to the context window. Understanding the capacity, compressibility, and decay of each layer is the basis for the discussions that follow.

\textbf{Three layers of carriers.} The innermost layer is the \emph{parameters}, whose capacity scales with model size and whose update cost is the highest, since it requires backpropagation and alignment data. Once written, however, parameters offer the most stable recall and represent capability that is genuinely internalized. The middle layer is \emph{external memory}, including vector stores, key-value caches, atomic memory items~\cite{xu2025amem}, and self-managed memory pools. This layer accumulates facts and experience without parameter updates, at low write cost and with high interpretability, but every use must pass through retrieval and its capacity is bounded by storage and retrieval bandwidth. The outermost layer is the \emph{context window}, which holds the system prompt, in-context examples, and tool-call history. It is the fastest channel accessible within a single inference pass, yet also the most volatile, because the window has a hard length limit and overly long contexts dilute attention. Across the three layers a clear trend emerges: as capability moves outward, write cost falls and interpretability rises, while recall stability and capacity ceilings fall in step.

\textbf{From ICL to TTT: the structural ceiling of context.} In-context learning, which operates entirely through context with parameters fixed, and test-time training~\cite{sun2024learning}, which updates part of the parameters at inference time, represent two ways of acquiring capability during inference. The migration from the former toward the latter is not incidental: it exposes the structural limit of a context-only route, in that information is discarded once it exceeds the window and any compression is necessarily lossy. This limit takes the form of two ceilings: a hard ceiling on length, since even windows extended to a million tokens or more remain bounded, and a soft ceiling on effective attention, since the well-documented needle-in-a-haystack phenomenon shows that usable attention over long contexts is unevenly distributed by position and effective information density does not grow linearly with window length. A practical observation reinforces this point: distilling rich tacit experience into a textual summary is brittle outside text-centric domains such as software engineering, where a compaction step can silently reverse a hard-won optimization because the rationale behind it never entered the summary~\cite{patel2025timelines}. Consequently, as task complexity grows and reasoning chains lengthen, context alone cannot carry full capability accumulation, and the update signal is forced back toward memory and parameters. The three layers are therefore not substitutes for one another but a hierarchy whose division of labor follows their timescales and capacity ceilings. A purely engineering route that tries to replace continual learning with ever longer context will meet a bottleneck on long-horizon tasks.

\textbf{Retrieval decay and the need for active forgetting.} Retrieval-augmented generation has long been read as a way to address continual learning through external memory, but as dialogue and task sequences lengthen it also decays: retrieval becomes dominated by stale documents, low-quality entries accumulate, relevance scores drift, and cross-session consistency degrades. This is the same motivation that led MemoryBank to introduce an Ebbinghaus-style forgetting curve~\cite{zhong2024memorybank}: external memory likewise requires an active forgetting mechanism. Recent surveys of agent memory frame its development as an evolution from storage to reflection to experience and identify long-range consistency and continual learning as its core drivers~\cite{luo2026storage}. The implication is that continual learning at the memory layer cannot focus only on how to write and retrieve, but must also study how to selectively reduce content, since otherwise the memory system degrades from a capability carrier into an accumulator of noise. The trade-off between stability and plasticity, classically posed at the parameter layer, thus shifts to the memory layer and calls for its own metrics.

Taken together, this section converges on a question that still lacks a systematic solution: which capabilities should be internalized into parameters over the long run, which should be retained in memory, and which should be used and discarded within context. Before asking how such scheduling might be organized, however, we first confront a prior objection: whether the harness engineering already used to manage these carriers by hand is in itself enough, which we take up next (Section~\ref{sec:outlook_engineering}).

\subsection{Why Harness-Level Accumulation Is Not Enough}
\label{sec:outlook_engineering}

Since no single carrier suffices, capability has to be spread across several at once, and something has to manage how they are combined. Today that management is largely done by hand, which prompts a question worth asking: whether the engineering progress now carried on the harness already amounts, in part, to continual learning. The question matters because many widely deployed means of extending capability, such as rule files in coding assistants, the evolution of system prompts in agent frameworks, and the steady growth of tool libraries, do not come from learning algorithms in the academic sense but from iterative engineering practice.

Our position is that the engineering progress on the harness already attains part of the goal of continual learning at the functional level, but does not replace it at the mechanistic level, for three reasons. \emph{First}, the harness offers externally editable capability rather than self-accumulated capability: a rule file or an agent tool list can be revised repeatedly by humans so that system behavior keeps improving, but the agent of that revision is a person, and it does not constitute a closed loop driven by the system's own experience. By continual learning we mean that the subject acquires and updates capability autonomously through interaction with the environment. \emph{Second}, the few methods that do close the loop, such as Promptbreeder~\cite{fernando2023promptbreeder}, Voyager~\cite{wang2023voyager}, and AgentEvolver~\cite{zhai2025agentevolver}, show that the harness layer can indeed carry continual learning in the genuine sense, yet their update mechanisms are either evolutionary search or reinforcement learning, and they remain far from mature in evaluation, stability, and reproducibility, as discussed in Section~\ref{sec:outlook_longhorizon}. \emph{Third}, capability on the harness is strongly coupled to context and transfers less readily than capability in parameters: a rule file written for one project or a skill library accumulated for one framework is bound to a specific context and is hard to transfer without loss when that context changes, whereas a parameter tensor, though costly to modify, carries capability that is comparatively transferable once internalized. A common conflation can now be clarified: the intuition that engineering progress is approaching AGI rests on the visible accumulation of capability on the harness, but a substantial part of that capability resides in the scaffolding around the model rather than in the model itself.

This connects to an open debate worth presenting evenhandedly. One position holds that continual learning remains essential even in the foundation-model era, because a deployed model is a snapshot of the world at training time and must contend with both task-shift and time-shift forgetting~\cite{bell2025futurecl}. A second position argues that the lack of continual learning is the primary bottleneck on the path from current systems to AGI, since prompt engineering and long rolling contexts are brittle patches that do not generalize beyond text-centric domains~\cite{patel2025timelines}. A third position counters that continual learning is a systems problem rather than a learning problem, contending that more context and more computation will let a system composed of several models, memory, and retrieval behave indistinguishably from one that learns continually, so that one need not make the model resemble a human too closely~\cite{lambert2025contra}. These positions map onto the central tension of this survey, between engineering compensation and mechanistic internalization. Our own reading is that engineering progress on the harness functionally substitutes for, but does not yet mechanistically constitute, continual learning, and that the role future continual-learning research should take on is to gradually convert the capability accumulation now carried by engineering practice into a process carried by the system's own mechanisms. This leaves an open question that motivates the rest of this section: if hand-tuned engineering is not in itself a self-driven mechanism, what might such a mechanism look like? We turn to that in the next section (Section~\ref{sec:outlook_coevolution}).

\subsection{Coordinating Memory, Skills, Protocols, and Parameters}
\label{sec:outlook_coevolution}

Having argued that hand-tuned engineering is functionally useful but not in itself a self-driven mechanism, we now ask what a genuine mechanism might look like. Neither model-only nor harness-only updates fully capture how capability is organized in LLM systems, since capability is now distributed across parameters, external memory, skills, and protocols at once. A natural direction is therefore to ask whether, and how, the model and the harness should be updated in a coordinated manner. Rather than treating the coordinated evolution of the two as a settled solution, we present it as a direction suggested by this multi-carrier organization of capability, and we frame it through three couplings that have not yet been studied systematically.

\textbf{Bidirectional transfer between parameters and harness.} The route taken by SKILL0, which explores skills on the harness before internalizing them into parameters~\cite{lu2026skill0}, illustrates one direction. The reverse direction may matter just as much: making latent capability that is hard to invoke from parameters explicit as a visible skill or memory item on the harness, so as to improve invocation efficiency and interpretability. A complete bidirectional mechanism would have to answer when to consolidate capability into parameters, for items that are frequently invoked and broadly shared across tasks, and when to return capability to the harness, for items that are context-specific or not yet stable. This has a structural analogy with cache and memory hierarchy management in computer systems, but no corresponding formal framework yet exists.

\textbf{Scheduling capability across carriers.} Seen this way, coordination is in large part a scheduling problem: what should stay in context, what should be written to memory, what should be consolidated into skills, and what should ultimately be internalized into parameters. Any such schedule would also have to account for the differing update cadences of the carriers, since memory is written most frequently, at the level of individual interactions, skills expand at an intermediate rate, at the level of tasks, and protocols and parameters are revised most slowly, at the level of projects or long-term constraints. Coordinating writing and reclamation across carriers that move at different rates is a key open problem for turning this direction from intuition into mechanism, and it too lacks a unified solution.

\textbf{Forgetting becomes multi-faceted and hard to localize.} In classical continual learning, forgetting is one-dimensional in that parameters are overwritten. Under composite carriers, it has at least four faces: catastrophic forgetting at the parameter layer, retrieval decay and entry aging at the memory layer, window overflow and attention dilution at the context layer, and capability mismatch at the skill and protocol layer caused by context drift. These differ in timescale, observability, and reversibility; they are hard to describe with a single metric and make it difficult to localize where forgetting actually occurs. Given this, it may be more useful to recast forgetting from a catastrophe at the parameter layer into a process of active compression and release across composite carriers, which would, in turn, call for unified criteria for which layer to act on, how fast, and what to retain or release. Developing such criteria is, on this view, the central obstacle to moving coordinated evolution from intuition to theory.


\subsection{Long-Horizon Agents as the Real Testbed}
\label{sec:outlook_longhorizon}

If coordinated updating along these lines is to be more than an intuition, it has to be testable. Future continual learning should be judged not by short tasks or static benchmarks, but by whether an agent can hold its goal, track state, correct errors, and accumulate experience across long, multi-step interactions in open environments. This is precisely the setting in which the carrier ceilings of Section~\ref{sec:outlook_carriers} and the coordinated updating discussed in Section~\ref{sec:outlook_coevolution} interact and where problems surface most readily.

\textbf{Why training on a fixed set is insufficient.} A static benchmark measures the capability a model ships with, whereas a long-horizon task measures the net gain of capability during operation. A model that scores highly on a fixed test set need not keep its goal from drifting or its state from being lost over tens or hundreds of interaction steps. This pushes the evaluation criterion from single-point accuracy toward trajectory-level measures, and it explains why inference-time update mechanisms such as test-time training become necessary: capability must be replenished during inference and deployment rather than frozen at training time. Recent long-horizon memory benchmarks and surveys respond to exactly this gap, casting agent memory as an evolution from storage to reflection to experience and treating long-range consistency as a primary driver~\cite{luo2026storage}.

\textbf{Memory compression and architecture are drawn in together.} Over long runs, memory cannot grow without bound and must be continually compressed, consolidated, and forgotten, echoing the selective reduction discussed in Section~\ref{sec:outlook_carriers}. When both context and external memory reach their ceilings, the burden of carrying long-horizon capability flows back to the model architecture itself. Long-horizon operation is therefore not merely a matter of longer tasks; it forces memory mechanisms, update timing, and architectural design to change in concert, which is also why long-horizon capability is rarely obtained through a single-point improvement.

\textbf{Error accumulation is the core difficulty.} When a task spans tens to hundreds of steps, small per-step errors accumulate and amplify along the call chain: an incorrect retrieval at step $k$ can lead the agent to invoke the wrong skill at step $k{+}1$ and shift later decisions as a whole. Such compounding error is studied in the sequential-learning setting of classical continual learning, but it takes on new features in LLM agent systems. First, errors arise not only at the parameter layer but across a composite state made up of memory, skills, protocols, and parameters. Second, most harness operations are non-differentiable, so conventional gradient-based sensitivity analysis no longer applies. Third, an agent lacks a clear boundary between training and evaluation, so an erroneous state may be written back into memory through immediate feedback and form a positive feedback loop. Reflexion corrects through verbal self-reflection~\cite{shinn2023reflexion} and TTRL corrects through deployment-time updates~\cite{ttrl2025}, showing that drift-resistant routes are feasible, yet both remain single-mechanism and lack a unified framework. Establishing a verifiable model of error accumulation and drift-resistant mechanisms for long-horizon agents is thus one of the most challenging and valuable directions for the coming years.

\subsection{Continual Learning as a Priority on the Path to AGI}
\label{sec:outlook_agi}

The most open-ended question is how far the memory-and-harness route still is from genuine AGI. Because any forecast of AGI risks excess optimism or pessimism, this subsection offers no definite answer. Instead, it draws together several signals visible in the developments reviewed above and then states a clear judgment about research priorities.

\textbf{Signal one: the unit of study extends from the model to the system.} Classical continual learning studies a model, whereas the real unit of work in the LLM era is increasingly a system composed of the model, memory, skills, protocols, tools, and environment. A paradigm worth pursuing is an end-to-end learnable agent framework, in which the components of the harness are no longer hand-built scaffolding but learnable modules trained and evolved jointly with the model. AgentEvolver offers an early demonstration~\cite{zhai2025agentevolver}, but a complete theory of system-level continual learning does not yet exist.

\textbf{Signal two: verifiable rewards as a bridge.} Reinforcement learning with verifiable rewards~\cite{shao2024deepseekmath}, training pipelines in the style of DeepSeek-R1~\cite{deepseekai2025deepseekr1}, and the use of verifiable signals at deployment time by TTRL~\cite{ttrl2025} point together to one shift: when a task outcome can be verified automatically, through code execution, the correctness of a mathematical solution, or passing unit tests, an agent can keep improving itself without human labels. Verifiable rewards are thus a key bridge from engineering progress to a continual-learning mechanism. The limitation is equally clear, since many real tasks, such as writing, consulting, and long-horizon planning, carry no explicit verifiable signal, so future work needs frameworks for semi-verifiable rewards or composite feedback.

\textbf{Signal three: forgetting reinterpreted as capability management rather than a defect.} As recast in Section~\ref{sec:outlook_coevolution}, selective forgetting is better seen as an active means of managing capability than as a passive decline; memory systems such as MemoryBank~\cite{zhong2024memorybank} and atomic memory~\cite{xu2025amem}, the pruning of tool libraries, and the finiteness of context all point this way, and the foundation-model perspective likewise treats selective forgetting as an important direction, distinguishing task-shift from time-shift forgetting~\cite{bell2025futurecl}.

\textbf{A judgment about AGI priorities.} We are inclined to locate the role of continual learning on the path to AGI as follows: continual learning for the frontier model is the first priority, whereas domain-specific continual learning is not. The value of future continual learning lies less in having a narrow model repeatedly absorb the knowledge of one domain, and more in letting the frontier model keep growing across a broad range of knowledge and skills. Achieving this requires coordinated improvement across these dimensions, spanning both the parameters and the harness, rather than relying on domain-specific knowledge alone or on single-point changes that touch only the model or only the harness, even though, at the present stage, changing only the model or only the harness can still yield sizable gains. This judgment is consistent with both poles of the debate above: it accepts the diagnosis that continual learning is a primary bottleneck toward AGI, and it also accepts the optimism that a systematized route can come close, while placing the emphasis on the systematic organization of capability at the level of the frontier model. It has further been argued that, once online learning is truly solved, a model could pool what it learns across all of its copies, so that a single system effectively learns every job, which helps explain why continual learning for the frontier model carries such an overriding priority~\cite{patel2025timelines}.

\textbf{An honest estimate of the distance.} Taking the signals together, we are inclined to believe that the memory-and-harness route is rapidly approaching the capability ceiling of a general assistant at the engineering level, but that genuine AGI, a system capable of autonomous capability expansion, cross-domain transfer, and robust long-horizon decision making, remains a considerable distance away, and that this distance will not be closed simply by longer context, larger memory, or more skills. Other conditions remain uncertain, yet continual learning is an unavoidable part of the path; recasting it from an isolated algorithmic problem into a systematic question of how capability is organized is the direction that the perspective advocated in this survey is meant to keep pointing toward.

\section{Conclusion}
\label{sec:conclusion}

his survey reframes continual learning in the LLM era as a boundary extension along three analytically distinct but interrelated dimensions: \When, \Where, and \How. Classical continual learning is situated as a specific point in the three-axis coordinate space, while LLM-era methods extend the boundary along one or more axes (Sections~\ref{sec:when}, \ref{sec:where}, and \ref{sec:how}) and increasingly depart from the classical coordinate on multiple axes at once, moving into the interior of the coordinate space (Section~\ref{sec:cross_axis_combinations}). The framework not only organizes the distribution of existing methods but, through the contrast between dense and empty regions, identifies unfilled coordinates that offer concrete directions for future work. We expect the taxonomy to serve as a unified reference for describing and advancing methods as LLM agent systems continue to mature toward engineering practice.

We acknowledge several limitations of this work. First, the three-axis taxonomy admits crossings at its boundaries; the classification of some cross-axis methods follows the primary-axis convention adopted in Section~\ref{sec:taxonomy_overview}, and alternative classifications are equally reasonable. Second, LLM continual learning is a rapidly iterating area; our coverage is limited to work publicly available before the submission deadline and may lag behind the most recent arXiv preprints. Third, evaluation for harness-layer and cross-axis methods is itself in rapid flux, so the evaluation gaps discussed in Section~\ref{sec:outlook_longhorizon} are necessarily a coarse summary, leaving the design of concrete protocols for future work.


\bibliographystyle{unsrtnat}
\bibliography{custom}

\clearpage
\appendix
\counterwithin{equation}{section}
\counterwithin{table}{section}
\counterwithin{figure}{section}

\section{Glossary of Abbreviations}
\label{app:glossary}

For readability, Table~\ref{tab:abbrev} summarizes the abbreviations used throughout this survey.

\begin{table}[h]
\caption{Abbreviations Used in This Survey.\label{tab:abbrev}}
\centering
\footnotesize
\renewcommand{\arraystretch}{1.1}
\begin{tabular}{@{}p{0.18\linewidth} p{0.72\linewidth}@{}}
\toprule
\textbf{Abbrev.} & \textbf{Expansion} \\
\midrule
CL          & Continual Learning \\
LLM         & Large Language Model \\
MLLM        & Multimodal Large Language Model \\
CPT         & Continual Pre-Training \\
ICL         & In-Context Learning \\
PEFT        & Parameter-Efficient Fine-Tuning \\
LoRA        & Low-Rank Adaptation \\
MoE         & Mixture of Experts \\
SFT         & Supervised Fine-Tuning \\
SDFT        & Self-Distillation Fine-Tuning \\
    RLHF        & Reinforcement Learning from Human Feedback \\
    RLVR        & Reinforcement Learning from Verifiable Rewards \\
    GRPO        & Group Relative Policy Optimization \\
    SDPO        & Self-Distillation Policy Optimization \\
    SRPO        & Sample-Routed Policy Optimization \\
    OPD         & On-Policy Distillation \\
OPSD        & On-Policy Self-Distillation \\
TTA         & Test-Time Adaptation \\
TTT         & Test-Time Training \\
TTRL        & Test-Time Reinforcement Learning \\
EWC         & Elastic Weight Consolidation \\
SI          & Synaptic Intelligence \\
MAS         & Memory Aware Synapses \\
LwF         & Learning without Forgetting \\
GEM         & Gradient Episodic Memory \\
OGD         & Orthogonal Gradient Descent \\
RAG         & Retrieval-Augmented Generation \\
KL          & Kullback--Leibler Divergence \\
\bottomrule
\end{tabular}
\end{table}

Table~\ref{tab:fig2-label-conventions} lists the Figure~\ref{fig:taxonomy_overview}
labels that require disambiguation or whose displayed short forms are not
explicit in the corresponding paper titles.

\begin{table}[htbp]
\caption{Selected short labels in Figure~\ref{fig:taxonomy_overview} and their corresponding references.}
\label{tab:fig2-label-conventions}
\centering
\footnotesize
\renewcommand{\arraystretch}{1.1}
\begin{tabularx}{\textwidth}{@{}p{0.23\textwidth}X@{}}
\toprule
\textbf{Figure 2 label} & \textbf{Corresponding paper} \\
\midrule
MER & \textit{Revisiting Replay and Gradient Alignment for Continual Pre-Training of Large Language Models}~\cite{abbes2026mer} \\
SimpleCPT & \textit{Simple and Scalable Strategies to Continually Pre-train Large Language Models}~\cite{ibrahim2024simple} \\
SLIM & \textit{SLIM: Let LLMs Learn More and Forget Less with Soft LoRA and Identity Mixture}~\cite{han2025slim} \\
SDFT & \textit{Self-Distillation Enables Continual Learning}~\cite{shenfeld2026sdft}; this is the on-policy SDFT shown in Figure~\ref{fig:taxonomy_overview} \\
RFT-CPT & \textit{Why Reinforcement Fine-Tuning Enables MLLMs Preserve Prior Knowledge Better: A Data Perspective}~\cite{zhang2025reinforcement} \\
SLAO & \textit{Merge before Forget: A Single LoRA Continual Learning via Continual Merging}~\cite{mergebeforeforget2025} \\
TMEM & \textit{Scaling Self-Evolving Agents via Parametric Memory}~\cite{ren2026parametric} \\
SELF-PARAM & \textit{Self-Updatable Large Language Models by Integrating Context into Model Parameters}~\cite{wang2024selfparam} \\
AgeMem & \textit{Agentic Memory: Learning Unified Long-Term and Short-Term Memory Management for Large Language Model Agents}~\cite{yu2026agemem} \\
MemoPilot & \textit{From Player to Master: Enhancing Test-Time Learning of LLM Agents via Reinforcement Learning over Memory}~\cite{cai2026memopilot} \\
AWM & \textit{Agent Workflow Memory}~\cite{wang2025awm} \\
ACE & \textit{Agentic Context Engineering: Evolving Contexts for Self-Improving Language Models}~\cite{zhang2026ace} \\
Dyn-Cheatsheet & \textit{Dynamic Cheatsheet: Test-Time Learning with Adaptive Memory}~\cite{suzgun2025dynamiccheatsheet} \\
\bottomrule
\end{tabularx}
\end{table}

\section{Benchmark Catalogue}
\label{app:benchmarks}

Table~\ref{tab:benchmark_catalogue} summarizes representative benchmarks and evaluation protocols used by the literature reviewed in this survey. The \emph{When}, \emph{Where}, and \emph{How} columns indicate the update regimes directly instantiated or evaluated by each benchmark; they are profile labels rather than mutually exclusive benchmark categories. The catalogue is representative rather than exhaustive.

\begingroup
\footnotesize
\setlength{\LTcapwidth}{\textwidth}
\setlength{\LTleft}{0pt}
\setlength{\LTright}{0pt}
\setlength{\tabcolsep}{2.5pt}
\newlength{\benchmarkcontentwidth}
\setlength{\benchmarkcontentwidth}{\dimexpr\textwidth-10\tabcolsep\relax}
\setlength{\emergencystretch}{1em}
\renewcommand{\arraystretch}{1.10}
\newcommand{\benchmarkentry}[2]{\textbf{#1}\newline\mbox{\cite{#2}}}
\begin{longtable}{@{}%
  >{\raggedright\arraybackslash}p{0.16\benchmarkcontentwidth}%
  >{\raggedright\arraybackslash}p{0.13\benchmarkcontentwidth}%
  >{\raggedright\arraybackslash}p{0.14\benchmarkcontentwidth}%
  >{\raggedright\arraybackslash}p{0.17\benchmarkcontentwidth}%
  >{\raggedright\arraybackslash}p{0.20\benchmarkcontentwidth}%
  >{\raggedright\arraybackslash}p{0.20\benchmarkcontentwidth}@{}}
\caption{Representative benchmarks and evaluation protocols viewed through the three analytical perspectives.}
\label{tab:benchmark_catalogue}\\
\toprule
\textbf{Benchmark / protocol} & \textbf{When} & \textbf{Where} & \textbf{How} & \textbf{Evaluation setting} & \textbf{Main outputs} \\
\midrule
\endfirsthead

\multicolumn{6}{@{}l}{\itshape Table~\ref{tab:benchmark_catalogue}, continued}\\
\toprule
\textbf{Benchmark / protocol} & \textbf{When} & \textbf{Where} & \textbf{How} & \textbf{Evaluation setting} & \textbf{Main outputs} \\
\midrule
\endhead

\midrule
\multicolumn{6}{r@{}}{\footnotesize Continued on next page}\\
\endfoot

\bottomrule
\endlastfoot

\benchmarkentry{TiC-LM}{li2025ticlm}
& Pre-training
& Parameters
& Off-policy
& Time-continual pre-training over 114 chronologically ordered Common Crawl dumps with time-stratified evaluation
& Adaptation to recent data, retention of earlier knowledge, held-out loss, and general-capability performance \\
\benchmarkentry{Temporal\allowbreak Wiki}{jang2022temporalwiki}
& Pre-training
& Parameters
& Off-policy
& Periodic language-model updates from differences between consecutive English Wikipedia snapshots, with Wikidata-based evaluation
& Acquisition of updated and new facts, retention of previous knowledge, perplexity, and update efficiency \\
\benchmarkentry{TRACE}{trace}
& Post-training
& Parameters
& Off-policy
& Sequential instruction tuning across eight heterogeneous text tasks
& Task performance, forgetting, general ability, and instruction following \\
\benchmarkentry{CITB}{zhang2023citb}
& Post-training
& Parameters
& Off-policy
& Continual instruction tuning on the InstrDialog and InstrDialog++ dialogue-task streams
& Average performance, forward and backward transfer, forgetting, and generalization to unseen tasks \\
\benchmarkentry{CoIN}{CoIN}
& Post-training
& Parameters
& Off-policy
& Multimodal continual instruction tuning over ten datasets from eight task categories
& Task performance, forgetting, instruction following, and general-knowledge retention \\
\benchmarkentry{MLLM-CL}{zhao2025mllmcl}
& Post-training
& Parameters
& Off-policy
& Multimodal continual learning under complementary domain-incremental and ability-incremental settings
& Domain and ability performance, average accuracy, forgetting, knowledge transfer, and update efficiency \\
\benchmarkentry{MLLM-CT\allowbreak Bench}{guo2025mllm}
& Post-training
& Parameters
& Off-policy; On-policy; Learning beyond gradients
& Continual instruction tuning across seven tasks and six domains, comparing supervised fine-tuning, reinforcement fine-tuning, and model fusion
& Final-answer accuracy, average performance, backward transfer, forgetting, and process-level reasoning quality \\
\benchmarkentry{CoTTA protocol}{wang2022cotta}
& Inference-time
& Parameters
& Off-policy
& Continual test-time adaptation on a non-stationary stream using teacher-based pseudo-labels
& Streaming error or accuracy, stability, error accumulation, and retention under domain shifts \\
\benchmarkentry{OAKS}{kim2026oaks}
& Inference-time
& Harness (Memory)
& Learning beyond gradients
& Fine-grained continual knowledge streams in OAKS-BABI and OAKS-Novel, where individual facts may change repeatedly
& State-tracking accuracy, adaptation delay, robustness to distraction, and memory-system reliability \\
\benchmarkentry{Memory\allowbreak Bench}{memorybench2025}
& Inference-time
& Harness (Memory)
& Learning beyond gradients
& Simulated continual user feedback across multiple domains, languages, and task types in deployed LLM systems
& Feedback utilization, adaptation and retention, task effectiveness, and computational efficiency \\
\benchmarkentry{EvoMem\allowbreak Bench}{evomembench2026}
& Inference-time
& Harness (Memory)
& Learning beyond gradients
& Standardized comparison of memory methods across in-episode versus cross-episode and knowledge- versus execution-oriented settings
& Task performance, long-context comparison, retrieval and execution effectiveness, and cross-setting robustness \\
\benchmarkentry{SkillLearn\allowbreak Bench}{skilllearnbench2026}
& Inference-time
& Harness (Skills)
& On-policy; Learning beyond gradients
& Iterative skill generation from agent experience on 20 verified tasks across 15 real-world sub-domains
& Skill quality, execution-trajectory quality, task success, iterative improvement, and recursive drift \\
\benchmarkentry{SkillFlow}{skillflow2026}
& Inference-time
& Harness (Skills)
& On-policy; Learning beyond gradients
& Lifelong skill discovery, patching, and reuse over 166 sequential tasks from 20 task families
& Task success, skill usage and utility, transfer, regression, and library evolution \\
\benchmarkentry{SEAGym}{seagym2026}
& Post-training; Inference-time
& Parameters; Harness (Protocols)
& Off-policy; On-policy; Learning beyond gradients
& Shared self-evolution protocol with train, frozen validation, held-out ID/OOD tests, replay diagnostics, snapshots, and cost records
& Held-out transfer, replay retention, evolution trajectories, update reliability, and cost \\
\end{longtable}
\endgroup

\section{Representative Cross-Dimensional Method Profiles}
\label{app:method-profiles}
\label{app:extended_table}

Table~\ref{tab:method-profiles} complements Fig.~\ref{fig:method_tree} by recording representative methods across the three perspectives. The entries are non-exclusive profile labels, not strict single-valued coordinates: a method may carry several labels within a perspective, and an arrow denotes a carrier trajectory over time. The table is not an exhaustive literature census, and its row counts should not be interpreted as estimates of research density.

\begingroup
\footnotesize
\setlength{\LTcapwidth}{\textwidth}
\setlength{\tabcolsep}{3.5pt}
\renewcommand{\arraystretch}{1.13}
\begin{longtable}{@{}%
  >{\raggedright\arraybackslash}p{0.16\textwidth}%
  >{\raggedright\arraybackslash}p{0.12\textwidth}%
  >{\raggedright\arraybackslash}p{0.21\textwidth}%
  >{\raggedright\arraybackslash}p{0.20\textwidth}%
  >{\raggedright\arraybackslash}p{0.25\textwidth}@{}}
\caption{Representative cross-dimensional method profiles. Semicolons indicate multiple applicable labels; arrows indicate a temporal carrier trajectory.}
\label{tab:method-profiles}
\label{tab:method_landscape}\\
\toprule
\textbf{Method} & \textbf{When} & \textbf{Where} & \textbf{How} & \textbf{Profile note} \\
\midrule
\endfirsthead

\multicolumn{5}{@{}l}{\itshape Table~\ref{tab:method-profiles}, continued}\\
\toprule
\textbf{Method} & \textbf{When} & \textbf{Where} & \textbf{How} & \textbf{Profile note} \\
\midrule
\endhead

\midrule
\multicolumn{5}{r@{}}{\footnotesize Continued on next page}\\
\endfoot

\bottomrule
\endlastfoot

CKL~\cite{jang2022towards}
& Pre-training
& Parameters
& Off-policy
& Sequential knowledge updating over evolving corpora \\
ELLE~\cite{qin2022elle}
& Pre-training
& Parameters
& Off-policy
& Expansion and function-preserving initialization during lifelong pre-training \\
TimeLMs~\cite{loureiro2022timelms}
& Pre-training
& Parameters
& Off-policy
& Diachronic language-model updates over temporal data slices \\
O-LoRA~\cite{wang2023orthogonal}
& Post-training
& Parameters
& Off-policy
& Orthogonal adapter subspaces mitigate interference across tasks \\
LoRAMoE~\cite{dou2024loramoe}
& Post-training
& Parameters
& Off-policy
& Modular LoRA experts preserve and route task-specific capability \\
SDFT~\cite{yang2024sdft}
& Post-training
& Parameters
& Off-policy
& Self-distillation uses a fixed teacher distribution during fine-tuning \\
COPR~\cite{zhang2025copr}
& Post-training
& Parameters
& Off-policy
& Preference learning regularizes successive policy updates \\
CPPO~\cite{zhang2024cppo}
& Post-training
& Parameters
& On-policy
& Continual reinforcement learning from human feedback \\
Self-distillation CL~\cite{shenfeld2026sdft}
& Post-training
& Parameters
& On-policy
& The current policy supplies responses used for continual self-distillation \\
AIMMerging~\cite{feng2025aimmerging}
& Post-training
& Parameters
& Learning beyond gradients
& Capability is combined through model merging \\
ZeroFlow~\cite{zeroflow2025}
& Post-training
& Parameters
& Learning beyond gradients
& Zeroth-order updates avoid ordinary back-propagated gradients \\
TTT-LM~\cite{sun2024learning}
& Inference-time
& Parameters
& Off-policy
& Self-supervised updates are performed on the observed test sequence \\
TTRL~\cite{ttrl2025}
& Inference-time
& Parameters
& On-policy
& Test-time reinforcement learning uses current-policy generations \\
A-MEM~\cite{xu2025amem}
& Inference-time
& Harness (Memory)
& Learning beyond gradients
& Agent memory is reorganized and written directly during use \\
Reflexion~\cite{shinn2023reflexion}
& Inference-time
& Harness (Memory)
& Learning beyond gradients
& Reflective text is generated and written to verbal memory \\
Voyager~\cite{wang2023voyager}
& Inference-time
& Harness (Skills)
& Learning beyond gradients
& Executable skills are accumulated in an external library \\
SkillRL~\cite{xia2026skillrl}
& Post-training
& Parameters; Harness (Skills)
& On-policy
& Reinforcement learning updates the policy while a skill library co-evolves \\
SkillWeaver~\cite{skillweaver2025}
& Inference-time
& Harness (Skills)
& Learning beyond gradients
& Web-agent skills are discovered, refined, and stored for reuse \\
Promptbreeder~\cite{fernando2023promptbreeder}
& Post-training
& Harness (Protocols)
& Learning beyond gradients
& Evolutionary search refines prompts and mutation prompts \\
GEPA~\cite{agrawal2025gepa}
& Post-training
& Harness (Protocols)
& Learning beyond gradients
& Reflective prompt evolution updates the surrounding protocol \\
AgentEvolver~\cite{zhai2025agentevolver}
& Post-training
& Parameters; Harness (Memory)
& On-policy
& Reinforcement learning updates policy parameters while selected trajectories accumulate in experience memory \\
SKILL0~\cite{lu2026skill0}
& Post-training
& Harness (Skills) $\rightarrow$ Parameters
& On-policy
& A curriculum first accumulates reusable skills and then internalizes selected capability \\
\end{longtable}
\endgroup

\end{document}